\documentclass[10pt,twocolumn,letterpaper]{article}

\usepackage{iccv}              
\usepackage{bbm}
\usepackage{multirow}
\usepackage{tabularx}
\usepackage{threeparttable}

\usepackage{enumitem}
\usepackage{listings}
\usepackage[T1]{fontenc}
\usepackage[scaled=.8]{beramono}   
\usepackage{soul}
\usepackage{xspace}
\usepackage{amsmath}
\sethlcolor{white}
\usepackage{mathtools}

\newcommand{\METHODNAME}{{\fontfamily{txtt}\selectfont {InfGen}}\xspace}
\newcommand{\TASKNAME}{{long-term traffic simulation}\xspace}
\newcommand{\TITLE}{Long-term Traffic Simulation with \\ Interleaved Autoregressive Motion and Scenario Generation}

\newcommand\mypar[1]{\par\noindent\textbf{#1}\;\;}
\definecolor{LightGrey}{rgb}{0.92,0.92,0.92}
\definecolor{Myred}{rgb}{1.00,0.12,0.36}
\definecolor{Myblue}{rgb}{0,0.60,0.87}
\definecolor{MotionBlue}{rgb}{210, 234, 254}
\definecolor{SceneGreen}{rgb}{220, 244, 221}
\definecolor{InitialAgent}{RGB}{206, 233, 252}
\definecolor{InitialAgentBoundary}{RGB}{226, 253, 255}
\definecolor{EgoAgent}{RGB}{232, 116, 142}
\definecolor{EgoAgentBoundary}{RGB}{252, 136, 162}
\definecolor{InsertedAgent}{RGB}{178, 234, 185}
\definecolor{InsertedAgentBoundary}{RGB}{198, 254, 205}
\usepackage{xspace}
\DeclareRobustCommand\onedot{\futurelet\@let@token\@onedot}

\usepackage{xcolor}

\definecolor{iccvblue}{rgb}{0.21,0.49,0.74}
\definecolor{motionblue}{RGB}{78,163,250}
\definecolor{scenegreen}{RGB}{117, 202, 112}
\definecolor{controlpurple}{RGB}{182, 160, 251}
\usepackage[pagebackref,breaklinks,colorlinks,allcolors=iccvblue]{hyperref}
\usepackage{booktabs}

\def\paperID{11456} 
\def\confName{ICCV}
\def\confYear{2025}

\title{\TITLE}

\author{Xiuyu Yang\footnotemark[1] \hspace{2em}
Shuhan Tan\footnotemark[1] \hspace{2em}
Philipp Kr\"ahenb\"uhl
\\
UT Austin\\
}

\begin{document}
\setlength{\fboxrule}{1pt}

\twocolumn[{
\renewcommand\twocolumn[1][]{#1}
\maketitle
\begin{center}
    \captionsetup{type=figure}
    \vspace{-1em}
    \includegraphics[width=\textwidth]{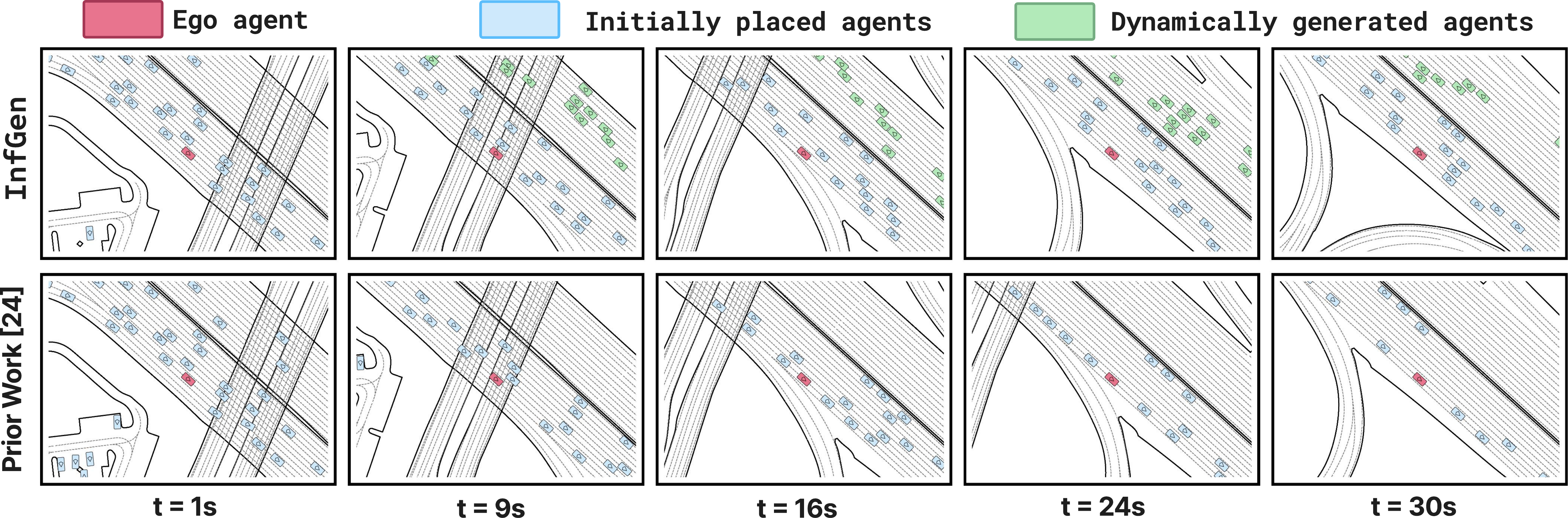}
    \captionof{figure}{Long-term traffic simulation with \METHODNAME and prior SOTA~\cite{nips24smart}. \METHODNAME keeps scene layout realistic while ~\cite{nips24smart} becomes empty.}

    \label{fig:teaser}
\end{center}
}]

\renewcommand{\thefootnote}{\fnsymbol{footnote}}
\footnotetext[1]{Equal contribution. Work done while Xiuyu interned at UT Austin.}
\renewcommand{\thefootnote}{\arabic{footnote}}

\begin{abstract}

An ideal traffic simulator replicates the realistic long-term point-to-point trip that a self-driving system experiences during deployment.
Prior models and benchmarks focus on closed-loop motion simulation for initial agents in a scene.
This is problematic for long-term simulation.
Agents enter and exit the scene as the ego vehicle enters new regions.
We propose~\METHODNAME, a unified next-token prediction model that performs interleaved closed-loop motion simulation and scene generation.
\METHODNAME automatically switches between closed-loop motion simulation and scene generation mode.
It enables stable long-term rollout simulation.
\METHODNAME performs at the state-of-the-art in short-term (9s) traffic simulation, and significantly outperforms all other methods in long-term (30s) simulation.
The code and model of \METHODNAME will be released at \href{https://orangesodahub.github.io/InfGen}{\textnormal{https://orangesodahub.github.io/InfGen.}}

\vspace{-5mm}
\end{abstract}
    
\section{Introduction}
\label{sec:intro}

Traffic simulation is a cornerstone of the extensive and safe development of self-driving systems.
The ultimate goal of traffic simulation is to create realistic trip-level driving experiences that faithfully reflect real-world self-driving conditions~\cite{ettinger2021womd, caesar2021nuplan, cusumano2025robust, wu2021flow, li2021metadrive}. 
A simulator should provide a realistic model of the environment, the ego-vehicle, and all other traffic agents throughout the trip.
Existing simulators 
easily handle an expansive static environment~\cite{ettinger2021womd, caesar2021nuplan, wu2021flow} and intricate ego-vehicle dynamics~\cite{cusumano2025robust, li2021metadrive}.
However, they often lack a stable long-term simulation of non-ego traffic agents.

In this paper, we introduce \METHODNAME, a long-term traffic simulator: Given a short (1 second) driving-log, \METHODNAME simulates realistic traffic flow around the ego agent for up to 30 seconds.
This long-term setting leads to new challenges.
The ego agent may move outside its initial simulation area, leading to logged agents moving out of sight and becoming irrelevant.
Furthermore, when the ego agents drive into new map area not covered in the log, these street areas have no agents.
Gradually, a scenario becomes sparser and eventually empty (\cref{fig:teaser} bottom row).
This is clearly unrealistic.
\METHODNAME models this by combining closed-loop \textit{motion simulation} with \textit{scene generation} to remove exiting agents and spawn new agents according to the spatial scene layout.

\begin{figure}[t]
    \centering
    \includegraphics[width=0.48\textwidth]{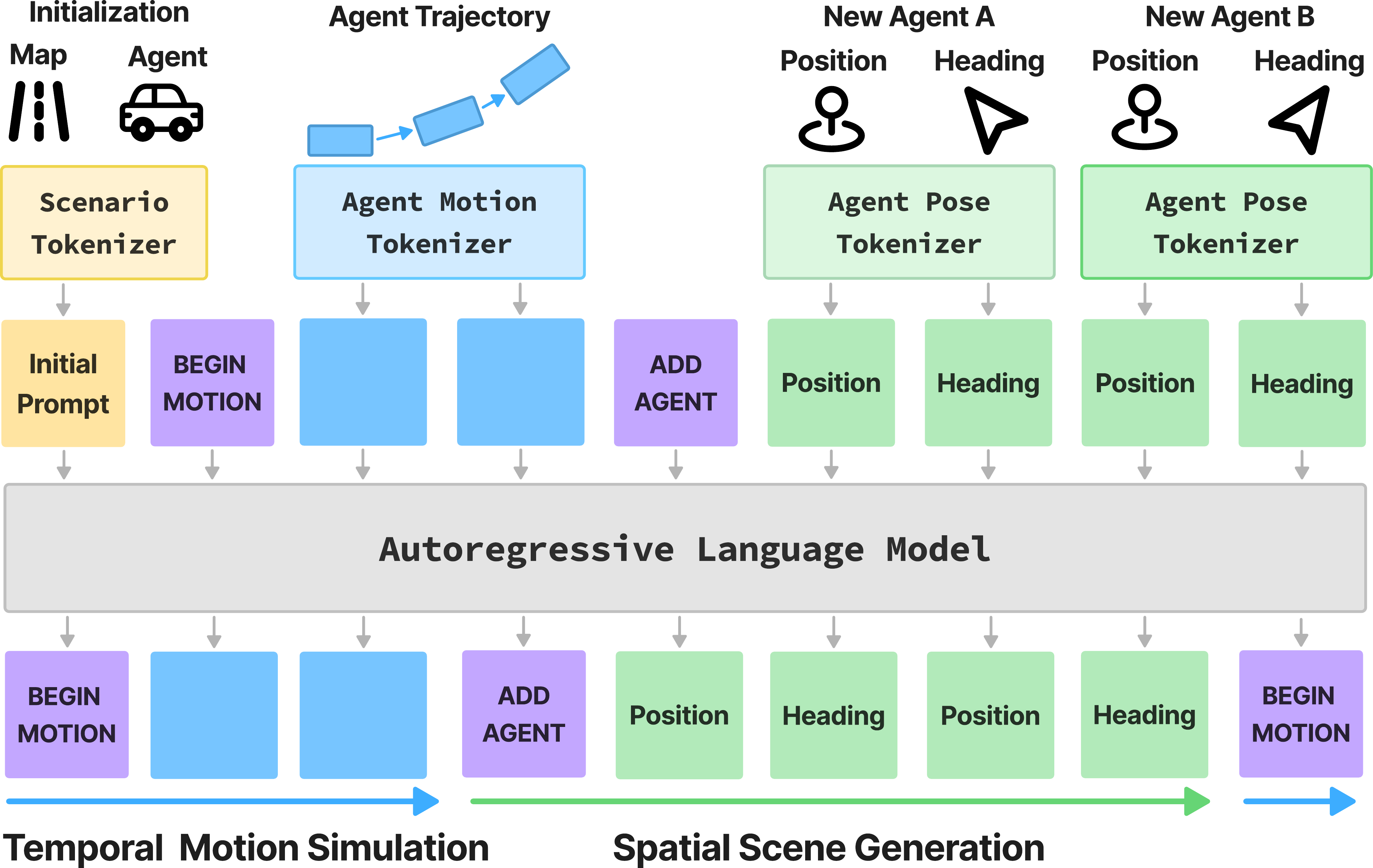}
    \caption{Overview of~\METHODNAME interleaved next-token-prediction process. Colors mark different token modalities.}
    \label{fig:infgen_pipeline}
    \vspace{-5pt}
\end{figure}

\begin{figure*}
    \centering
    \includegraphics[width=\textwidth]{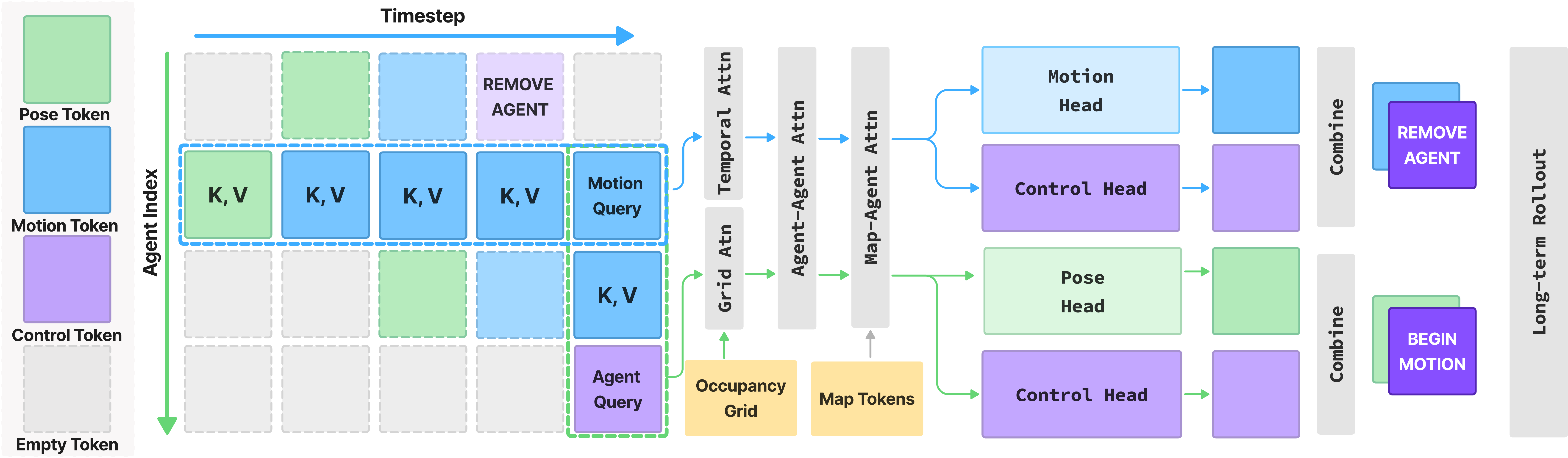}
    \caption{Pipeline of~\METHODNAME interleaved motion simulation (\textcolor{motionblue}{\text{blue flow}}) and scene generation (\textcolor{scenegreen}{\text{green flow}}). For either task, we first pass its query feature through blocks of attention layers and feed it to a task-specific head and a control head.
    We then sample from both heads to obtain a \textcolor{motionblue}{\text{motion token}} or \textcolor{scenegreen}{\text{pose token}}, as well as a \textcolor{controlpurple}{\text{control token}}, which determines determines which task to execute next.
    }
    \vspace{-5pt}
    \label{fig:model}
\end{figure*}

\METHODNAME (\cref{fig:infgen_pipeline}) is a unified autoregressive transformer with interleaved token prediction.
It handles \textit{temporal} motion simulation and \textit{spatial} scene generation in a unified model.
We design a set of tokenizers to convert task-specific behaviors of motion simulation and scenario generation into discrete tokens.
We then add mode-control tokens to mark the task switch between the two tasks, indicating what the current task is and when to switch.
This design allows us to convert each real log into a single ordered sequence of tokens containing interleaved data of both tasks.
We directly train~\METHODNAME with the next token prediction objective end-to-end on real data.
Noticeably, thanks to the next token prediction formulation, we can train~\METHODNAME on short-term driving logs and produce stable long-term rollouts, up to $6\times$ longer than the training horizon (Fig.~\ref{fig:teaser} top).
We show a detailed pipeline of~\METHODNAME in Fig.~\ref{fig:model}.

We show in Section~\ref{sec:exps} that \METHODNAME significantly outperforms prior SOTA models~\cite{nips24smart, zhang2024catk} in 30-second \TASKNAME for in terms of both motion and scenario realism, showing its strong capacity for stable long-horizon rollout.
We contrast different models' rollout with visualizations (Fig.~\ref{fig:rollouts}).
Furthermore, \METHODNAME achieves strong performance even on the standard short-term Sim Agent setting~\cite{montali2023waymo}, showing its strong adaptivity to different rollout horizons.
In addition, we provide a comprehensive set of analysis on the \TASKNAME task to shed light on research towards trip-level driving simulators.

\section{Related Work}
\label{sec:related}

\vspace{2mm}

\mypar{Closed-loop Motion Simulation.}
In traffic simulators, motion simulation aims to model realistic multi-agent interactions that mimic real-world data.
Early works like Wayformer~\cite{nayakanti2023wayformer} focus on open-loop imitation learning from real logs~\cite{Xu_2023, salzmann2020trajectron++,zhao2019multi, zhang2024trafficbots, yang2023rmmdet}.
Other works like ProSim~\cite{tan2024promptable} and CAT-K~\cite{zhang2024catk} instead focus on modeling closed-loop interaction between agents~\cite{casas2020implicit, suo2021trafficsim, zhang2023learning}.
CtRL-Sim~\cite{rowe2024ctrl} learns reactive agents by applying offline reinforcement learning to diverse traffic scenarios.
More recent works like SMART~\cite{nips24smart} discover the effectiveness of modeling this task as an autoregressive next-token-prediction problem~\cite{philion2023trajeglish, hu2024gump, zhao2024kigras, zhou2024behaviorgpt, seff2023motionlm}.
ProSim~\cite{tan2024promptable} enables multimodal prompts to control behavior semantics of any agent.
Most recently, GIGAFLOW~\cite{cusumano2025robust} shows strong agent performance emerges from large-scale self-play in simulation.
For this direction, nuPlan~\cite{caesar2021nuplan} and WOSAC~\cite{montali2023waymo} provide data and benchmark for fair comparisons.
All these works focus on simulating motions of agents existed in the history, leading to unrealistic scene layouts under long-term rollout.
\METHODNAME solves this issue with interleaved scene generation, maintaining realistic scene layout across the rollout horizon.

\vspace{2mm}
\mypar{Traffic Scenario Generation.} 
This line of work focuses on generation realistic and interesting traffic scenarios.
Early works like SceneGen~\cite{tan_2021_scenegen} and TrafficGen~\cite{trafficgen} generates agent initial poses on an empty map.
Another popular direction is to generate near-collision scenarios with adversarial optimization~\cite{rempe2022generating, wang2021advsim, xie2024advdiffuser, xu2023diffscene, chang2024safesim}.
More recent works like LCTGen~\cite{tan2023language} enables better customizations of the generated scenarios in forms of text~\cite{tan2023language, zhong2023language}, scenario queries~\cite{ding2023realgen} or cost functions~\cite{zhong2023guided}.
SLEDGE~\cite{chitta2024sledge} combine generative models with rule-based traffic simulation to synthesize dynamic driving scenarios.
These works either focus on static scene layout initialization, or only generate short-term open-loop scenarios.
In contract, \METHODNAME conducts dynamic scenario layout generation during closed-loop rollout, enabling stable long-term simulation.
Concurrent works like SceneDiffuser++~\cite{Tan_2025_CVPR} and ScenarioDreamer~\cite{rowe2025scenario} introduces a vectorized latent diffusion approach to generate realistic and diverse driving simulation environments.

\vspace{2mm}
\mypar{Interleaved Next-Token Prediction.}
Recent advances in vision-language models have sparked works that unify generation and understanding tasks with interleaved mixed-modal token sequences.
For example, Chameleon~\cite{Chameleon_2024} proposes to train LLMs on interleaved text and image tokens in any arbitrary sequences. This line of works show strong performance on traditional multimodal tasks, but also long-form mixed modal generation that interleaves between image and text generation~\cite{Chameleon_2024, tian2024mminterleaved, xie2024showo, Kou2024Orthus, xu2025modality}.
\METHODNAME follows the same philosophy but focus on a different pair of modalities: temporal agent motion and spatial agent layout.

\section{Problem Formulation} 
\label{sec:problem}

The goal of traditional traffic simulation~\cite{montali2023waymo} is to predict future agent trajectories given historical observations (with time span $T_\text{H}$) and a static map.
Specifically, at timestep $t_0$, we are given the static map $\mathcal{M}$ and the history states of all the agents $\mathcal{A}_{0:t_0} = \{a_{0:t_0}^1, a_{0:t_0}^2, \dots, a_{0:t_0}^N\}$, where each agent $a^i$ has history states up to timestep $t_0$.
The standard task is to predict future agent states over a fixed horizon $T$, formulated as estimating the conditional distribution: $p(\mathcal{A}_{t_0+1:T} \mid \mathcal{M}, \mathcal{A}_{0:t_0})$.
Prior work~\cite{nips24smart, tan2024promptable} factorize the simulation on the time axis and turn it to an autoregressive prediction task:
\begin{equation}
p(\mathcal{A}_{t_0+1:T} \mid \mathcal{M}, \mathcal{A}_{0:t_0}) = \prod_{t = t_0}^{T-1} p(\mathcal{A}_{t+1} \mid \mathcal{M}, \mathcal{A}_{0:t}).
\end{equation}
However, this formulation assumes a fixed set of agents throughout the prediction horizon, which does not hold true for long-term simulations with the horizon $T^{\prime} ( \gg T )$.
In realistic scenarios (\textit{e.g.}, ~\cite{montali2023waymo}), agents dynamically enter and leave the observable region around the ego vehicle.
To address this issue, we model two interleaved processes at each timestep: 1) motion simulation: predicts future motions for existing agents; 2) scene generation: dynamically inserting new agents and removing agents exiting the scenario.
At each timestep $t$, we first performs motion simulation: $\mathcal{A}_{t+1} \sim p_{\text{motion}}(\mathcal{A}_{t+1} \mid \mathcal{M}, \mathcal{A}^{\prime}_{0:t})$, followed by scene generation: $\mathcal{A}^{\prime}_{t+1} \sim p_{\text{scene}}(\mathcal{A}^{\prime}_{t+1} \mid \mathcal{M}, \mathcal{A}_{t+1})$.
Here, $\mathcal{A}_{t+1}$ contains the predicted motions of existing agents, while $\mathcal{A}^{\prime}_{t+1}$ represents the updated set of agents after adding new agents and removing exiting agents. 
Then, we formulate the \textit{long-term traffic simulation} task as:
\begin{equation}
\label{eq:formulation}
\begin{aligned}
    p(\mathcal{A}^{\prime}_{t+1:T^{\prime}} \mid \mathcal{M}, \mathcal{A}_{0:t_0}) 
    = \prod_{t = t_0}^{T^{\prime}-1}  &p_{\text{scene}}(\mathcal{A}^{\prime}_{t+1} \mid \mathcal{M}, \mathcal{A}_{t+1})\times\\
    & p_{\text{motion}}(\mathcal{A}_{t+1} \mid \mathcal{M}, \mathcal{A}^{\prime}_{0:t}).
\end{aligned}
\end{equation}

\section{\METHODNAME}
\label{sec:method}


\METHODNAME is a unified autoregressive model for long-term traffic simulation.
Regarding the inputs, it first tokenizes all scene context information ($\mathcal{M}$ and $\mathcal{A}_{0:t_0}$) into sequences of discrete tokens, see \cref{sec:tokens}.
It then uses an autoregressive model for interleaved next-token prediction, see \cref{sec:interleave}.
And \cref{sec:modelarch,sec:train} provide the details on the model architecture and training process of~\METHODNAME.



\subsection{Tokenization}
\label{sec:tokens}
We aim to convert all agent motions, layouts and the map in a real log into a sequence of discrete tokens.
We tokenize each modality differently.

\vspace{2mm}
\mypar{Map Tokenizer.} 
We adapt the map tokenizer from~\cite{nips24smart}: we uniformly segment all the road elements into a set of fixed-length road vectors.
Each vector contains the corresponding features, including start/end points, directions and road type.
We collect all road vectors to the map token set $\mathcal{V}_{\text{map}}$.



\vspace{2mm}
\mypar{Agent Motion Tokenizer.} 
We follow the motion tokenizer from prior works~\cite{nips24smart,zhang2024catk}.\kern-0.1em
\footnote{Please refer to~\cite{nips24smart} for the details of motion tokenizer and k-disks approach.}
Specifically, we segment the continuous trajectories of all agents in the dataset with a fixed time span of 0.5 seconds.
Then we use k-disks algorithm~\cite{nips24smart} to cluster these trajectories into the set of motion vocabulary $\mathcal{V}_{\text{motion}}$.
Finally, at every 0.5-second interval, we convert the continuous trajectory into a discrete token with the index of its nearest neighbor in the motion vocabulary.


\vspace{2mm}
\mypar{Agent Pose Tokenizer.} 
When a new agent is inserted into the scenario, its initial pose (position and heading) is given.
We tokenize the pose of each inserted agent as a pair of discretized \textit{position} and \textit{heading} tokens.
For position, we construct a grid with a radius of $R$, centered on the ego agent's location with x-axis aligned with ego agent's heading, resulting in position token set $\mathcal{V}_{\text{pos}}$.
To get the position token, we obtain the index of the grid closest to the agent based on L2 distance. 
For heading, we divide the $360^{\circ}$ range at the interval of $\Delta \theta$, resulting in heading token set $\mathcal{V}_{\text{head}}$.
To get the heading token, we similarly obtain the index of the heading interval closest to the agent heading.
For simplicity, we refer the pair of position and heading tokens as a \textit{pose token}.


\vspace{2mm}
\mypar{Mode Control Tokenizer.}
We model long-term traffic simulation as interleaved motion simulation and scenario generation.
We design the control tokens $\mathcal{V}_\text{control}$, consists of 4 special tokens, to mark mode transition between tasks:
1) \texttt{<BEGIN MOTION>}: the next token is an agent motion token to simulate current existing agents; 2) \texttt{<ADD AGENT>}: the next token is an agent pose token to insert a new agent; 3) \texttt{<KEEP AGENT>}: the current agent is kept in the scenario; 4) \texttt{<REMOVE AGENT>}: the current agent will be removed in the next timestep.
In the next section we will show how to use $\mathcal{V}_\text{control}$ to control the interleaved simulation process.


\vspace{2mm}
Tokenizing a real log into a discrete token sequence allows us to convert the complex mixture-task simulation process into a simple interleaved next token prediction task.

\subsection{Interleaved Next Token Prediction}
\label{sec:interleave}

\vspace{2mm}
\mypar{Dynamic Agent Matrix.} 
Traffic simulation can be represented by an agent matrix shown in Figure~\ref{fig:model}.
The horizontal axis represent \textit{temporal} lifecycle of each agent: being inserted, active moving and finally exit the scenario.
The length of the temporal axis equals to the rollout horizon.
On the other hand, the vertical axis represent \textit{spatial} agent layout at each timestep, where the width represent the number of active agents at each step.
When a new agent is inserted, a new row is created and append to the matrix.
Conversely, when a current agent gets removed, its row gets deleted from the matrix.
For long-term scenarios, because agents are frequently being inserted and removed from the scenario, the number of rows of the matrix are also constantly changing.
Hence, we term it the \textit{dynamic} agent matrix.

As shown in Figure~\ref{fig:model}, we represent the long-term traffic simulation task as extending the dynamic agent matrix on different axis.
Motion simulation (the upper blue flow) extends the temporal axis by adding new columns with predicted motion tokens.
In contrast, scenario generation (the lower green flow) extends the spatial axis by adding new rows with pose tokens of new agents, and removing current rows of exiting agents.
The control tokens determine how to interleave these two processes.

\vspace{2mm}
\mypar{Temporal Motion Simulation.}
We show this process in the blue flow of Figure~\ref{fig:model}.
For the $i_{\text{th}}$ active agent at timestep $t$, we use its current motion token $m^t_i$ as the query $q_{m^t_i}$ to obtain its motion feature $f_m$\kern-0.1em
\footnote{We omit the $t$ and $i$ in this section when possible for simplicity.}.
Specifically, we input $q_m$ to a Temporal Attention layer to attend to the key and value of all its own past motion tokens within $t_\text{w}$ timesteps (within the same row as the query token):
\begin{equation}
\label{eq:temp_attn}
    q^{\prime}_{m^t}=\text{{\fontfamily{pcr}\selectfont MHSA}}^\text{t}(q_{m^t}, \{k_{m^{t-\tau}}\}_{\tau=1}^{t_\text{w}}, \{v_{m^{t-\tau}}\}_{\tau=1}^{t_\text{w}}).
\end{equation}
The output is then sent to an Agent-Agent Attention layer to attend to all the other $N^t$ active agents within a valid range $r^{\text{a}\leftrightarrow \text{a}}$ at the same timestep $t$:
\begin{equation}
\label{eq:agent-agent}
    q^{\prime\prime}_{m_i}=\text{{\fontfamily{pcr}\selectfont MHCA}}^{\text{a}\leftrightarrow \text{a}}(q_{m_i}^{\prime}, \{k_{m_j}\}_{j=1}^{N^\text{t}}, \{v_{m_j}\}_{j=1}^{N^\text{t}}).
\end{equation}
Finally, the query goes through a Map-Agent Attention layer to attend to the $N_r$ precomputed map tokens within a valid range $r^{\text{m}\leftrightarrow \text{a}}$:
\begin{equation}
    f_{m_i} = \text{{\fontfamily{pcr}\selectfont MHCA}}^{\text{m}\leftrightarrow \text{a}}(q_{m_i}^{\prime\prime}, \{k_{m_j}\}_{j=1}^{N_\text{r}}, \{v_{m_j}\}_{j=1}^{N_\text{r}}).
\label{eq:map-agent}
\end{equation}

The motion head and control head separately take $f_m$ and output the probabilities over the motion and control tokens, from which we sample a motion token and a control token for each active agent.
In this subtask, we enforce the control token to be sampled from \texttt{<KEEP AGENT>} and \texttt{<REMOVE AGENT>}.
If control token is \texttt{<KEEP AGENT>}, we add the sampled motion tokens to the next column of each agent.
Otherwise, if the control token is \texttt{<REMOVE AGENT>}, we add this control token to the next column and discard the motion token.
The above process is conducted for all the current active agents \textit{in parallel} in training and inference.
After this, we switch to the scene generation step.

\vspace{2mm}
\mypar{Spatial Scene Generation.}
We show this process in the green flow of Figure~\ref{fig:model}.
After each motion simulation step, we use a learnable agent query $a_0$ to obtain the scene generation feature $f_{a_0}$.
Same as the motion query, the agent query is also sent through three attention layers to collect the context information.
The latter two layers are the same as the motion query, while the Temporal Attention layer is replaced by a Grid Attention layer.
This layer allows the agent query to attend to the occupancy grid tokens $g$ of total size $N_\text{g}=N_\text{p}=\left|\mathcal{V_\text{pos}}\right|$ derived from the position tokens:
\begin{equation}
    q^{\prime}_{a_0} = \text{{\fontfamily{pcr}\selectfont MHCA}}^\text{g}(q_{a_0}, \Gamma(\{k_{g_j}\}_{j=1}^{N_\text{g}}), \Gamma(\{v_{g_j}\}_{j=1}^{N_\text{g}})),
\end{equation}
where $\Gamma(\cdot)$ presents the transformation preceding the attention calculation for efficiency. We then pass $q^{\prime}_{a_0}$ through the other two layers (Eq.~\ref{eq:agent-agent} and Eq.~\ref{eq:map-agent}) to produce $f_{a_0}$. 
Differently, $q^{\prime}_{a_0}$ will have various visible range for the active agents $r^{q\leftarrow a}$ at current timestep and for the map tokens $r^{q\leftarrow m}$, please refer to Appendix~\ref{sec:supp_model} for more details.

Then the pose head and control head take $f_{a_0}$ and output distributions for pose and control tokens respectively, from which we sample a pose token and a control token for each generation step
(Please refer to Appendix~\ref{sec:supp_model} for more details).
We enforce the control token to be sampled from \texttt{<ADD AGENT>} and \texttt{<BEGIN MOTION>} for this subtask.
Controlled by \texttt{<ADD AGENT>}, we append a new row to the agent matrix and assign the sampled pose token to the current timestep.
Then, conditioned on the all active agents, including the newly inserted one, we repeat the above step to autoregressively insert another new agents.
This process is terminated when the sampled control token is \texttt{<BEGIN MOTION>}.
In this case, we end the scene generation process and move to motion simulation of the next timestep.
Finally, we remove any row that has a \texttt{<REMOVE AGENT>} token at the current timestep from the agent matrix.

\subsection{Model Architecture}
\label{sec:modelarch}

\vspace{2mm}
\mypar{Token Embedding.} 
Our model takes tokens of different modalities.
To model them with a single token sequence, we take different MLP layers to embed different kinds of tokens into the same latent dimension $D$ before entering the model. Please refer to Appendix~\ref{sec:supp_model} for more details.
\vspace{2mm}
\mypar{Modeling Layer.}
Our transformer model is composed of $L$ blocks of attention layers.
As mentioned in Section~\ref{sec:interleave}, each block contains 4 attention layers: Temporal Attention, Agent-Agent Attention, Map-Agent Attention and Grid Attention.
Here the first layer are implemented with multi-head self attention layer ($\text{{\fontfamily{pcr}\selectfont MHSA}}$), while the other layers are multi-head cross attention layer ($\text{{\fontfamily{pcr}\selectfont MHCA}}$).
Furthermore, we apply the position-aware attention from prior works~\cite{nips24smart, tan2024promptable} to explicitly model the relative positions between tokens.
Please refer to the Appendix for the details.

\vspace{2mm}
\mypar{Occupancy Grid Encoder.} We obtain occupancy grid features $f_\text{g}$ of the current scenario with the agent position tokens and map tokens via $\Gamma(\cdot)$.
Specifically, given the vocabulary size $N_\text{p}$ of position tokens $\mathcal{V}_\text{pos}$, we directly assign each token with occupation indication $\{0,1\}$, leading to an agents occupancy grid $g^{1\times N_\text{p}}\in \{0,1\}$.
We utilize the occupancy maps of agents in decoding pose token process, to make the agent query efficiently infer the spatial distributions of agents. We use an MLP Layer to convert $g^{1\times N_\text{p}}$ to its features $f_\text{g}^{N_\text{p}\times D}$ before feed it into Grid Attention layers.




\subsection{Training}
\label{sec:train}

\vspace{2mm}
\mypar{Ground-truth Sequence.}
As described in Section~\ref{sec:problem}, real-world traffic scenario log~\cite{montali2023waymo} naturally contains data of agent motion, insertion and removal behaviors.
To train our model with the interleaved NTP problem explained in Section~\ref{sec:interleave}, we convert each ground-truth log into an ordered sequence of token labels.
To this end, we enforce a specific ordering to chain tokens from different modalities.
Specifically, at each timestep we arrange each type of tokens with a fixed order: 1) motion tokens from all the current agents; 2) control tokens \texttt{<REMOVE AGENT>} and \texttt{<KEEP AGENT>} for the current agents; 3) pose tokens and \texttt{<ADD AGENT>} for any inserted agents; 4) \texttt{<BEGIN MOTION>} that mark the transition to the next timestep.
For the same type of tokens we order them following to the agent's distance to the ego agent from near to far.
With this rule we obtain a sequence of ground-truth (GT) token labels from each real log to train our model.
Please refer to Appendix~\ref{sec:supp_token} for more details about the GT tokens.

\vspace{2mm}
\mypar{Learning Objective.} 
As shown in Figure~\ref{fig:infgen_pipeline}, given the GT input tokens as input, our model predicts the distributions of different kinds of tokens at the corresponding position. 
We then train our model with a set of standard NTP objective for each type of tokens.
For example, the loss function for the motion token is:
\begin{equation}
\begin{aligned}
    \mathcal{L}_{\text{motion}} = - \sum_{t=1}^{T-1} \log{p_\theta(m_i^{t+1}\mid c^{1:t})},
\label{eq:loss_motion}
\end{aligned}
\end{equation}
where $T$ is the total number of timesteps,  $m_i^{t+1}$ is the GT motion token in the next timestep, $p_\theta(m_i^{t+1}|c^{1:t})$ is the model-predicted probability of the GT token. 
Here, $c^{1:t}$ is the ensemble of all history context the model attends to, including history motions of the same agent, positions of other agents in the current timestep, and the map.
We formulate the loss function in the same way for agent pose tokens $\mathcal{L}_{\text{pose}}$ and mode control tokens $\mathcal{L}_{\text{control}}$.
We also have supervision of shapes and types for those new agents.
Our total training loss can be written as:
\begin{equation}
\begin{aligned}
\mathcal{L} = \lambda_1\mathcal{L}_{\text{motion}} + \mathcal{L}_{\text{pose}} + \lambda_4\mathcal{L}_{\text{control}} + \mathcal{L}_\text{attr},
\label{eq:loss}
\end{aligned}
\end{equation}
where $\mathcal{L}_{\text{pose}}=\lambda_2\mathcal{L}_{\text{pos}} + \lambda_3\mathcal{L}_{\text{head}}$ and $\mathcal{L}_{\text{attr}}=\lambda_5\mathcal{L}_{\text{shape}} + \lambda_6\mathcal{L}_{\text{type}}$. We directly end-to-end train~\METHODNAME with $\mathcal{L}$ on the fully tokenized dataset.
During training~\METHODNAME not only learns how to conduct the two tasks respectively, but also learns to automatically and seamlessly switch between them.
Please refer to Appendix~\ref{sec:supp_train} for more training explanations.


\section{Experiments}
\label{sec:exps}
In this section, we validate our work from two aspects: \textit{1) How does \METHODNAME compare to the SOTA baselines on conventional short-term traffic simulation task in standard benchmarks? 2) How does \METHODNAME perform on our \textit{mainly} introduced \TASKNAME task?}

\begin{table}[t]
    \centering
    \small
    \raggedleft
    \caption{Short-term traffic simulation in WOSAC~\cite{montali2023waymo} ($\uparrow$).}
    \renewcommand{\arraystretch}{1.0}
    \setlength{\tabcolsep}{2.5pt}
    \begin{tabularx}{\linewidth}{l>{\centering\arraybackslash}X>{\centering\arraybackslash}X>{\centering\arraybackslash}X>{\centering\arraybackslash}X}
        \toprule[1pt]
        Method & Composite & Kinematic & Interactive & Map \\
        \midrule[0.5pt]
        TrafficBots~\cite{zhang2024trafficbots} & 0.6976 & 0.3994 & 0.7103 & 0.8342 \\
        GUMP~\cite{hu2024gump} & 0.7404 & 0.4773 & 0.7872 & 0.8339 \\
        SMART-7M~\cite{nips24smart} & 0.7521 & 0.4799 & 0.8048 & 0.8573 \\
        CatK~\cite{zhang2024catk} & 0.7603 & 0.4611 & 0.8103 & 0.8732 \\
        \METHODNAME & 0.7514 & 0.4754 & 0.7936 & 0.8502 \\
        \bottomrule[1pt]
    \end{tabularx}
    \vspace{-10pt}
    \label{table:short}
\end{table}

\renewcommand{\arraystretch}{1.0}
\begin{table*}[ht]
    \centering
    \caption{Long-term traffic simulation evaluation ($\uparrow$) on WOMD validation split measured by the metrics introduced in Sec.~\ref{sec:metric}.
    }
    \begin{tabular}{ccccc cc c c c}
        \toprule[1pt]
        \multirow{2}{*}{Method} & \multirow{2}{*}{Composite} & \multirow{2}{*}{Kinematic} & \multirow{2}{*}{Interactive} & \multirow{2}{*}{Map-based} & \multicolumn{5}{c}{Placement-based} \\
        & & & & & overall & $N_+$ & $N_-$ & $D_+$ & $D_-$ \\
        \midrule[0.5pt]
         SMART-7M~\cite{nips24smart} & 0.6519 & 0.5839 & 0.7542 & 0.8102 & 0.4324 & 0.5713 & 0.4964 & 0.3371 & 0.3248 \\
         CatK~\cite{zhang2024catk} & 0.6584 & 0.5850 & 0.7584 & 0.8186 & 0.4424 & 0.5842 & 0.5233 & 0.3371 & 0.3248 \\
         \METHODNAME & 0.6606 & 0.5966 & 0.7619 & 0.8087& 0.4542 & 0.6273 & 0.5635 & 0.3169 & 0.3092 \\
        \bottomrule[1pt]
    \end{tabular}
    \label{table:newmetric}
    \vspace{-1em}
\end{table*}

\vspace{2mm}
\mypar{Dataset.} We train and validate \METHODNAME on Waymo Open Motion Dataset (WOMD)~\cite{ettinger2021womd}, with $\sim480\mathrm{K}$ scenarios for training and $\sim44\mathrm{K}$ scenarios for validation. Each scenario consists of a $T=9.1\,\mathrm{s}$ recorded rollout, with the first $T_\text{H}=1.1\,\mathrm{s}$ historical frames and the subsequent $8\,\mathrm{s}$ future frames.

\vspace{2mm}
\mypar{Training.} To train \METHODNAME, we use a total batch size of 8 on 8 NVIDIA A5000 GPUs with the AdamW optimizer and cosine annealing learning rate scheduler. The initial learning rate is 0.0005. For the loss function in Eq.~\ref{eq:loss}, we set $\lambda_1=\lambda_3=1$, $\lambda_2=\lambda_4=10$, $\lambda_5=0.2$, $\lambda_6=5$.

\subsection{WOMD Sim Agent Challenge}
We first evaluate \METHODNAME on the standard \textit{short-term} traffic simulation task in WOSAC~\cite{montali2023waymo}.
For this setting, we enforce~\METHODNAME to skip all the scene generation steps, and predict the rollout for only $8s$.
We then evaluate the rollouts under the official WOSAC metrics and compare the results with the top-performing methods in Table.~\ref{table:short}.
The results show \METHODNAME performs very competitive even under the short-term setting without any task-specific tuning, achieving similar performance to SOTA~\cite{zhang2024catk} and outperforms strong models~\cite{hu2024gump, zhang2024trafficbots}.

\subsection{Long-term Traffic Simulation Setup}
\label{sec:metric}

\vspace{2mm}
\mypar{Rollout Setup.} 
To be compatible with the map size of WOMD, we set the total rollout duration of our long-term traffic simulation experiments as \( T^{\prime}=31.1\,\mathrm{s} \) with \( \mathrm{FPS}=10 \).
The first \( 1.1\,\mathrm{s} \) corresponds to the historical segment, from which we simulate 300 future steps.


\vspace{2mm}
\mypar{WOSAC Metric Adaption.} 
The standard WOSAC metric~\cite{montali2023waymo} evaluates short-term traffic simulation by comparing simulated rollouts directly with ground-truth (GT) logs over an 8-second horizon. Specifically, it computes 9 metrics that assess aspects such as kinematic realism, interaction realism, and map adherence, assuming a one-to-one correspondence between simulated and logged agents.

However, in our long-term simulation setting, the rollout duration extends significantly beyond the standard 8-second window, reaching up to 30 seconds. This creates two critical challenges: (1) there is no direct one-to-one correspondence between simulated agents and logged agents over the entire long-term rollout, as agents dynamically enter and exit the scene; (2) the evaluation window (8 seconds in WOSAC) is shorter than our simulation horizon (30 seconds).
Therefore, we need to adapt the original WOSAC metric to suit our long-term simulation setting.

\begin{figure}[t]
    \centering
    \includegraphics[width=0.48\textwidth]{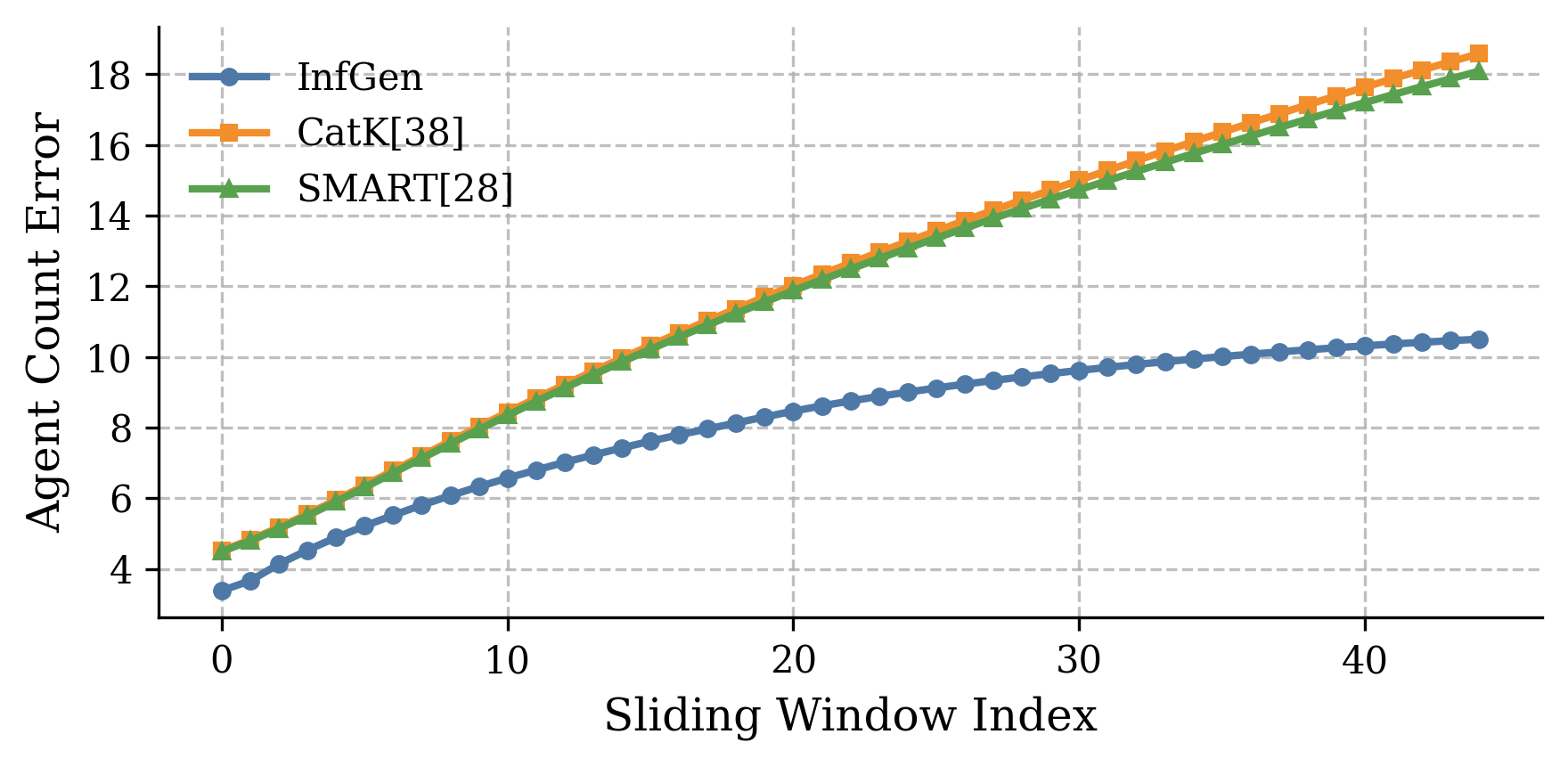}
    \caption{Agent Count Error (ACE) curves of \METHODNAME against baselines over 30s long-term simulation rollouts.}
    \label{fig:exp_ace}
    \vspace{-15pt}
\end{figure}

\begin{figure*}
    \centering
    \includegraphics[width=\textwidth]{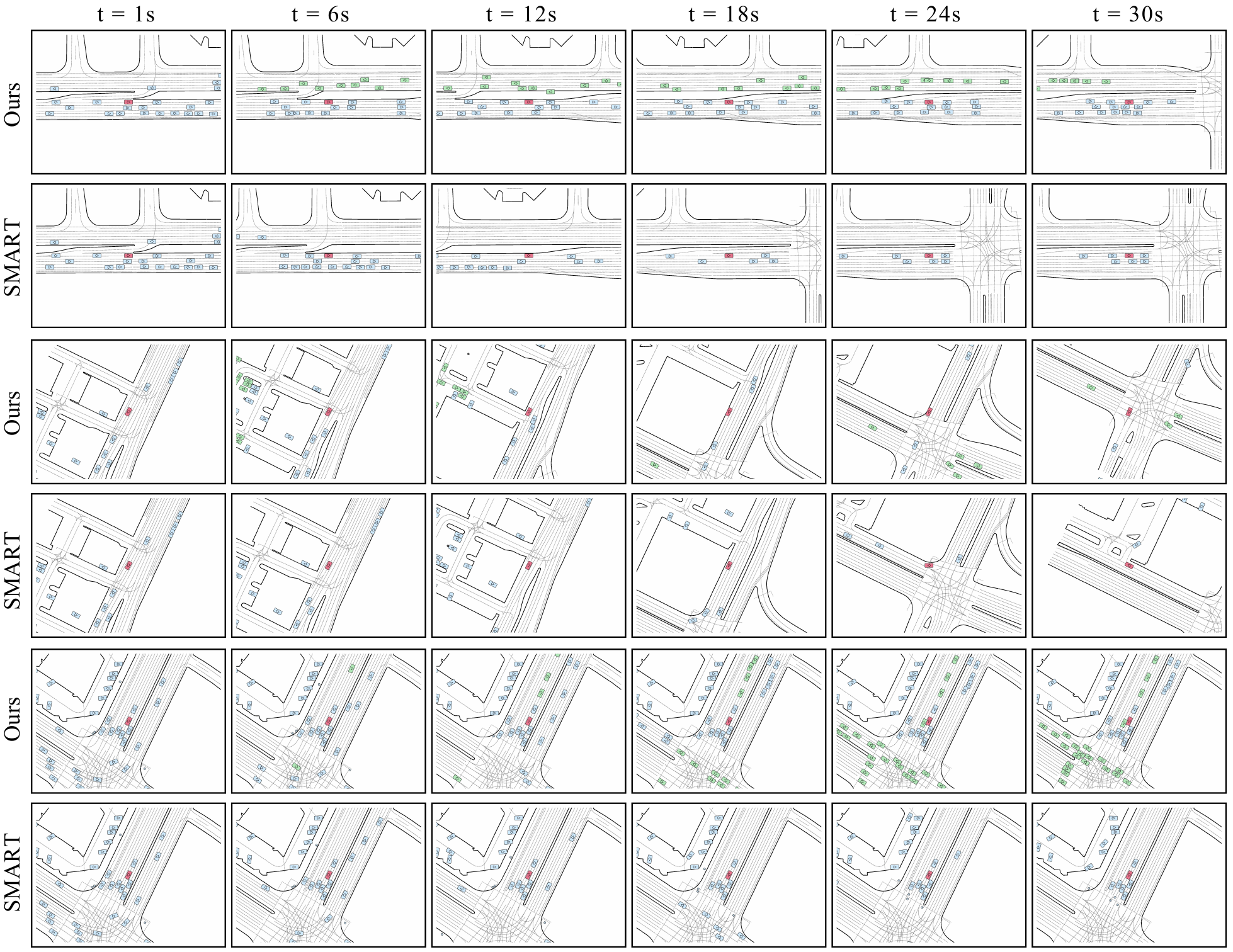}
    \vspace{-2em}
    \caption{\textbf{Qualitative results of long-term closed-loop rollouts for 5 scenarios.} We compare rollouts of \METHODNAME and SMART~\cite{nips24smart} here.
    \fcolorbox{InitialAgent}{InitialAgentBoundary}{\makebox[1pt]{\phantom{\rule{0pt}{1pt}}}} are the initially placed agents, \fcolorbox{InsertedAgent}{InsertedAgentBoundary}{\makebox[1pt]{\phantom{\rule{0pt}{1pt}}}} are the new agents inserted by \METHODNAME, and \fcolorbox{EgoAgent}{EgoAgentBoundary}{\makebox[1pt]{\phantom{\rule{0pt}{1pt}}}} are the ego agents. Please refer to Supplementary Materials for more results.}
    \label{fig:rollouts}
\end{figure*}

Specifically, we adapt the WOSAC evaluation as follows.
Given a long-term simulated rollout $\sigma^{\prime}=(\mathcal{M}, \mathcal{A}_{0\,:\,T^{\prime}}^{\prime})$, we extract short segments using a sliding window approach. Specifically, we slide a window of length $T_\mathrm{w} = T - T_\mathrm{H}$ (matching the standard $8\,\mathrm{s}$ evaluation window) at a fixed interval $\Delta t$ throughout the entire simulated rollout. Each sliding window generates a short-term segment $\mathcal{A}_{t\,:\,t+T_\mathrm{w}}^{\prime}$ that matches the length of standard evaluation segments.
Finally, we get $\mathcal{A}_{0:T^{\prime}}^{\prime}=\{\mathcal{A}_{\Delta s(i-1)\,:\,\Delta s(i-1)+T_\mathrm{w}}^{\prime}\}_{i=1}^P$ with the number of segments equal to $P$, and $\Delta s(P-1)+T_\mathrm{w}=T^{\prime}$.

Since the number of simulated agents $N^{\prime}$ in these segments may differ from the logged agents $N$, we cannot directly apply the original WOSAC evaluation. Instead, we compute agent-level Negative Log-Likelihood (NLL) scores for all simulated agents by evaluating their behaviors against a global distribution learned from the entire validation dataset ($\sim$ 48K scenarios). Concretely, we first estimate empirical distributions for agent motions, interactions, and placements from the entire validation dataset, and then measure how well our simulated agent behaviors conform to these reference distributions. This modified approach ensures a fair and consistent evaluation of long-term simulation realism with varying numbers of agents and rollout durations.

\vspace{2mm}
\mypar{Placement-based Metrics.} 
The standard WOSAC metrics evaluate simulation realism through multiple components: \textit{kinematic-based}, \textit{interaction-based}, and \textit{map-based}. These metrics are then aggregated into an overall realism composite metric using predefined weights. However, these metrics assume a fixed set of agents throughout the evaluation horizon and thus fail to capture the realism of agent insertion and removal events that are essential in long-term traffic simulations.
We then propose the \textit{placement-based} component, which comprises 4 types of statistics: the number of placement $N_+$, the number of removal $N_-$, the distance of placement $D_+$ and the distance of removal $D_-$, where the distances here are relative to the ego agent.
We aim to use the placement-based metrics to assess the realism of \METHODNAME in modeling the entry and exit of agents during the long-term rollout.
Similarly, we calculate the NLLs of the placement-based statistics under the logged distributions in the same way as the other components.

\vspace{2mm}
\mypar{Agent Count Error Metrics.} In addition to WOSAC metrics extensions, we also introduce a new Agent Count Error (ACE) metric to evaluate scene realism during long-horizon rollouts. For each sliding window over the $30\,\text{s}$ rollout, we compute the mean absolute difference in agent count between simulation and the ground-truth distribution of the validation set. We summarize the results with two scalar metrics (lower is better): \textit{Mean ACE} (overall error) and \textit{ACE Slope} (error growth rate via linear regression).

\vspace{2mm}
With the above setup we are now ready to evaluate different methods for long-term traffic simulation.

\subsection{Long-term Traffic Simulation Evaluation}
\label{sec:exp_long_term}

\vspace{2mm}
\mypar{Baselines.} Since existing works do not focus on long-term rollout, to fairly compare, our baselines are derived by improving upon the SOTA simulation methods \textbf{SMART}~\cite{nips24smart} and \textbf{CatK}~\cite{zhang2024catk}. SMART leverages vectorized map and agent trajectory data, predicting motion sequences through a decoder-only transformer architecture. CatK further introduces a closed-loop supervised fine-tuning technique and achieve better WOSAC simulation performance.

First, we extend their rollout duration to $T^{\prime}=31.1s$ and then calculate the kinematic, interactive and map-based metrics. For the placement-based metrics, we design a heuristic approach: we obtain the entered and exited agents by partitioning the distance between agents and the ego agent by the radius $R$, which corresponds to the \textit{distance placement} and \textit{distance removal} of placement-based measurements, introduced in Sec.~\ref{sec:metric}.
During the rollout process, we have the distances from each agent to the ego agent $\{\mathcal{D}^{\,i}_{0\,:\,T^{\prime}}\}_{i=1}^N$. At step $t$, the agents that first run within the range $R$, whose $\mathcal{D}_t \leq R$ given $\mathcal{D}_{0\,:\,t-1}> R$, are considered as \textit{entered} agents, while those agents with $\mathcal{D}_{t}>R$ and $\mathcal{D}_{0\,:\,t-1}\leq R$, are considered as \textit{exited} agents. Throughout the experiment, we adjust $R$ to achieve the highest placement-based score for baselines on validation set. In this case, we assign additional validity values to each agent at each timestep, thus only those valid agents will be included in baseline inferences and evaluation metrics. We follow the default settings in their papers.

\vspace{2mm}
\mypar{Quantitative Results.} Under the settings and proposed evaluation metrics in Sec.~\ref{sec:metric}, Table.~\ref{table:newmetric} shows the results of \TASKNAME under extended WOSAC metrics.
As can be seen, \METHODNAME achieves better realism performance than baselines. For placement-based metrics, our method significantly outperforms the baselines, demonstrating that our approach effectively models the spatial sequences of agents.
The lower scores on map-based metrics may be attributed to the insertion of agents in regions outside driving lanes or near road boundaries, leading to less realistic motion trajectories.

Figure~\ref{fig:exp_ace} shows the curve of comparison results under our introduced ACE metrics, where our method significantly outperforms prior methods: \METHODNAME achieves Mean ACE of 8.1 while baselines (CatK~\cite{zhang2024catk},SMART~\cite{nips24smart}) have scores of 12.2 and 12.0, which reflects the improvements of our dynamically scenario generation; \METHODNAME has ACE Slope: 0.15, however, baselines accumulate such errors with slopes of 0.32 and 0.31. This significantly confirms the much better long-term stability observed in Figure~\ref{fig:rollouts}.

\begin{table}[t]
    \centering
    \small
    \raggedleft
    \caption{Long-term motion prediction evaluation ($\uparrow$).}
    \renewcommand{\arraystretch}{1.0}
    \setlength{\tabcolsep}{2.5pt}
    \begin{tabularx}{\linewidth}{l>{\centering\arraybackslash}X>{\centering\arraybackslash}X>{\centering\arraybackslash}X>{\centering\arraybackslash}X}
        \toprule[1pt]
        Method & Composite & Kinematic & Interactive & Map \\
        \midrule[0.5pt]
        SMART-7M~\cite{nips24smart} & 0.7428 & 0.5413 & 0.7626 & 0.8349 \\
        CatK~\cite{zhang2024catk} & 0.7316 & 0.5216 & 0.7347 & 0.8495 \\
        \METHODNAME & 0.7432 & 0.5495 & 0.7685 & 0.8213 \\
        \bottomrule[1pt]
    \end{tabularx}
    \label{table:long2}
    \vspace{-10pt}
\end{table}


\vspace{2mm}
\mypar{Qualitative Analysis.} We further show \METHODNAME through the visualizations of the long-term rollouts. As depicted in Figure.~\ref{fig:rollouts}, the results highlight two core properties of \METHODNAME: \textit{long-term} and \textit{closed-loop}.
The first scenario depicts a bidirectional driving road, where \METHODNAME successfully simulates oncoming traffic. In the third scenario, when the rollout reaches $t=18s$, most of active agents run outside the scenario, resulting in an empty region, as reflected by prior works.
Under the control of \METHODNAME, new agents enter the scenario, allowing continued agent-agent and agent-map interactions with high realism.
While in baselines, those regions around the ego agent remains empty. 
Moreover, from the forth scenario, we observe our model can place new agents not only on driving roads but also in parking lots or other open areas.

\vspace{2mm}
\mypar{Motion-only Analysis.}
In Section~\ref{sec:problem} we formulate the long-term traffic simulation task as a product of motion simulation and scene generation (Equation~\ref{eq:formulation}).
Here we investigate the performance of different models in long-term traffic simulation when we only consider motion simulation.
Specifically, we disable agent insertion and removal for all the methods, and simply let them rollout for 30 seconds.
We then evaluate the rollouts with the adapted WOSAC metrics.
As shown in Table~\ref{table:long2}, all the compared methods have very similar performance.
This indicate that simply extending rollout horizon will not reveal important aspects of long-term traffic simulation, like the empty scenarios we see for SMART rollouts in Figure~\ref{fig:rollouts}.


\vspace{2mm}
\mypar{More Experiments and Analysis.}
Additional experiments and results are provided in Sec.~\ref{sec:addtional_results} of the Supplementary Materials, including further analysis on long-term traffic simulation, as well as ablation studies on our various tokens. Please refer to it for more details.
\vspace{-10pt}

\section{Conclusion}
\label{sec:conclusion}
In this work we propose \METHODNAME, a unified next-token prediction model for long-term traffic simulation. 
\METHODNAME learns to automatically switch between temporal motion simulation and spatial scene generation.
Our experiments show \METHODNAME significantly outperforms prior methods in long-term simulation, while keeping strong performance on standard short-term simulation.
We believe \METHODNAME is a steady step towards realistic trip-level traffic simulation.

\vspace{2mm}
\mypar{Limitations and Future Works.} The main limitation of this paper is that we did not evaluate \METHODNAME under a real trip-level rollout duration (> 5 minutes).
Our main constraint is the map areas available in WOSAC scenarios are too small: even in our 30s rollout the ego agent often drive to areas where no map is available.
We plan to further explore with suitable data.
Another limitation is that \METHODNAME is trained purely with supervised learning on logged real-world data, which may lead to failures due to overfitting and the model capturing spurious causal relationships.
Given the interactive nature of this task and its autoregressive generation process, in future work we plan to explore interactive reinforcement learning to further encourage realistic agent interactions and scenario generation through environment feedback~\cite{tan2025interactiveposttrainingvisionlanguageactionmodels, chen2025rift}.


{
    \small
    \bibliographystyle{ieeenat_fullname}
    \bibliography{main}
}

\renewcommand{\thesection}{\Alph{section}}
\clearpage
\appendix
\maketitlesupplementary

Our supplementary contains following contents:

\begin{enumerate}[label=(\Alph*), align=left]
    \item \textbf{Demo Video.} We provide more illustrative videos to demonstrate the motivation and demos in Sec.~\ref{sec:supp_demo}.
    \item \textbf{Model Details.} In addition to the key components introduced, we describe other modules of \METHODNAME in Sec.~\ref{sec:supp_model}.
    \item \textbf{Token Details.} We have a comprehensive demonstration of tokens design in \METHODNAME in Sec.~\ref{sec:supp_token}.
    \item \textbf{Training Details.} We explain how we train the interleaved autoregressive model, \METHODNAME, in Sec.~\ref{sec:supp_train}. 
    \item \textbf{Additional Results.} We have more experiments of long-term simulations and ablations in Sec.~\ref{sec:supp_more_results}.
    \item \textbf{Limitations and Future Direction.} We analyze the failure cases in \METHODNAME. And further discuss the limitations and the potential improvements in Sec~\ref{sec:fail}.
    \item \textbf{License.} We list licenses of all assets used in \METHODNAME. 
\end{enumerate}

\section{Demo Video}
\label{sec:supp_demo}

We provide additional videos for better demonstration of \METHODNAME. In this video, we showcase the existing problem of current baselines, as introduced in Sec.~\ref{sec:intro}, and the comparison between baselines and our method. Then we have more qualitative examples. Please refer to our \href{https://orangesodahub.github.io/InfGen/}{project page} for details.

\section{Model Details}
\label{sec:supp_model}

In this section, we provide more details of \METHODNAME model. We give an overview of the main hyperparameters of the model architecture, and explain the setup of agent embedding and query which are directly operated by transformer decoder. Then we describe more about how to decode various outputs.

\vspace{2mm}
\mypar{Hyperparameters.} We extend the base model from SMART-7M~\cite{nips24smart} to train out \METHODNAME model. We also have the detailed descriptions about the transformer decoders in Sec.~\ref{sec:interleave}, and we list all the main hyperparameters of the model architecture and our implementations in Table.~\ref{tab:supp_model}.

\begin{table}[!ht]
\renewcommand{\arraystretch}{0.95}
\centering
\caption{Main hyperparameters of \METHODNAME model.}
\vspace{-5pt}
\label{tab:supp_model}
\setlength{\tabcolsep}{3.5pt}
\begin{tabular}{l >{\centering\arraybackslash}p{2cm}}
    \toprule[1.5pt]
    Hyperparameter & Value \\
    \midrule[1pt]
    \multicolumn{1}{l}{\textit{Transformer Decoder}} & \\
    attention head dimension & 16 \\
    number of attention heads & 8 \\
    number of motion transformer blocks & 6 \\
    number of scene transformer blocks & 3 \\
    number of map transformer blocks & 3 \\
    $D$, feature dimension of token embedding & 128 \\
    number of frequency bands & 64 \\
    $t_\text{w}$, Agent temporal Attention radius & 12 \\
    $r^{\text{a}\leftrightarrow\text{a}}$, Agent-Agent Attention radius & 60 \\
    $r^{\text{m}\leftrightarrow\text{a}}$, Map-Agent Attention radius & 30 \\
    $r^{\text{m}\leftrightarrow\text{m}}$, Map-Map Attention radius & 10 \\
    $r^{\text{q}\leftarrow\text{a}}$, Query-Agent Attention radius & 10 \\
    $r^{\text{q}\leftarrow\text{m}}$, Query-Map Attention radius & 75,10 \\
    \midrule[0.5pt]
    \multicolumn{1}{l}{\textit{Tokenizer}} & \\
    $R$, radius of position grids & 75 \\
    $\Delta g$, grid interval of $\mathcal{V}_\text{pos}$ & 3 \\
    $\Delta \theta$, angle interval of $\mathcal{V}_\text{head}$ & 3 \\
    $\left|\mathcal{V}_\text{motion}\right|$, vocabulary size of motion tokens & 2048 \\
    $\left|\mathcal{V}_\text{map}\right|$, vocabulary size of map tokens & 1024 \\
    $\left|\mathcal{V}_\text{pos}\right|$, vocabulary size of position tokens & 1849 \\
    $\left|\mathcal{V}_\text{head}\right|$, vocabulary size of heading tokens & 120 \\
    $\left|\mathcal{V}_\text{control}\right|$, vocabulary size of control tokens & 4 \\
    \bottomrule[1.5pt]
\end{tabular}
\end{table}

Note that we also inherit the road network from~\cite{nips24smart} (See Sec.3.3 of their paper) in our NTP pre-training process (where the Map-Map Attention Layers are incorporated). Since we adopt the map tokenizer and transform maps into discrete tokens, this network will help the model to understand the topological connectivity and continuity of unordered map tokens. Finally, the total model size of \METHODNAME is around 11M.

\subsection{Agent Feature Learning}

\vspace{2mm}
\mypar{Agent Embedding.} In our stacked attention layers, we directly operate on different tokens according to each specific task.
For practical purposes, we construct a comprehensive aggregated agent embedding $F_\text{agent}\in \mathbb{R}^{T\times D}$ (where $T$ denotes the rollout horizon) to process them uniformly as token sequences in the transformer layers, as mentioned in Sec.~\ref{sec:modelarch}.
Specifically, we obtain the $k_m$, $v_m$ in Eq.~\ref{eq:temp_attn},~\ref{eq:agent-agent},\,\ref{eq:map-agent} from tokens of different modalities through direct fusion, as Eq.~\ref{eq:agent_embed}. We first concatenate all the features on channel dimension and then produce the agent embedding through 1-layer MLP which is directly operated by transformer layers:
\begin{equation}
\label{eq:agent_embed}
    \text{{\fontfamily{pcr}\selectfont MLP}}(\text{{\fontfamily{pcr}\selectfont Concat}}(F_\text{motion},F_\text{position},F_\text{validity},F_\text{attribute})),
\end{equation}
where \textit{validity} denotes if the agent at current timestep is visible by the environment, which is labeled in real log data. And $F_\text{attribute}$ aggregates the shape and type values of each agent. Notably, $\textit{validity}\in\mathbb{R}^{T\times1}$ here reflects not only the state of agent but also implicitly embeds the control tokens.
To get all these features from their low-dimension values, we use MLP Layers for those in continuous space (\textit{e.g.}, agent shapes) and learnable embeddings for those in discrete space. Finally, we have the dynamic agent matrix tensor $F_{\mathcal{A}^\prime}\in\mathbb{R}^{A\times T\times D}$ (as the matrix in Fig.~\ref{fig:model}) which is continuously updated during the rollout in an interleaved autoregressive manner.

\vspace{2mm}
\mypar{Agent Query.} As we explained in Sec.~\ref{sec:interleave}, we start from an agent query $a_0$ to autoregressively insert the agents in spatial scene generation.
Ideally, we consider that the spatiotemporal features of the agent query are fully aligned with the ego agent, \textit{i.e.}, it follows the ego agent's position and motion. Since we are primarily concerned with the environment around the ego agent in this task, and the position tokens $\mathcal{V}_\text{pos}$ are also centered at the ego agent.
To build this query, we take exactly the same approach as Eq.~\ref{eq:agent_embed}. 
For its motion token, we directly use another new special token instead of any existing token from $\mathcal{V}_\text{motion}$.
For its position token, we fix it as the centroid of the $\mathcal{V}_\text{pos}$.
For its validity, we set it as invalid. And we use new special agent type and shape values as its inherent attributes.

\subsection{Modeling Layer}

\vspace{2mm}
\mypar{Position-aware Attention.} We explicitly model the relative spatial-temporal positions between input tokens in attention calculation (as in Eq.~\ref{eq:temp_attn},~\ref{eq:agent-agent},\,\ref{eq:map-agent}). For each query-context token pair (\textit{e.g.}, agent-agent or agent-map), we add the relative positional encoding from context token $F_{\text{c}}$ to query token $F_{\text{q}}$ ($F_{\text{c}},F_\text{q}\in\mathbb{R}^{D}$ are from $F_{\mathcal{A}^\prime}$).

Specifically, we incorporate 3 types of descriptors: relative distance $\Delta p$, the relative direction $\Delta d$, the relative heading $\Delta\theta$, where $p\in \mathbb{R}^2$ and $\theta\in (-\pi,\pi)$ denotes the positions and headings of input tokens.
For temporal attention module (Eq.~\ref{eq:temp_attn}), we add additional time span $\Delta t$ to formulate 4D descriptors:
\begin{equation}
\begin{aligned}
\label{eq:rel_pos}
    \Delta p_\text{cq}&=||p_\text{c}-p_\text{q}||_2, \; \Delta\theta_\text{cq}=\theta_\text{c}-\theta_\text{q}, \; \Delta t_\text{cq}=t_\text{c}-t_\text{q},\\
    \Delta d_\text{cq}&=\text{atan2}(\text{p}_{\text{c},\text{y}}-p_{\text{q},\text{y}},\;p_{\text{c},\text{x}}-p_{\text{q},\text{x}})-\theta_\text{q},
\end{aligned}
\end{equation}
which are formulated to $r_{ij}\in\mathbb{R}^D$, the relative positional encodings of tokens $F_{i}$,$F_{j}$ and then added to keys, values in geometric attention layers (Eq.\ref{eq:temp_attn},\ref{eq:agent-agent},\ref{eq:map-agent}):
\begin{align}
    r_{ij}&=\text{{\fontfamily{pcr}\selectfont PE}}([\Delta p_{ij},\Delta d_{ij},\Delta\theta_{ij},\Delta t_{ij}]), \label{eq:pos_enc} \\
    F_{\mathcal{A}^\prime}^{l}&=\text{{\fontfamily{pcr}\selectfont Attn}}(\phi_q(F_{\mathcal{A}^\prime}^{l-1}),\phi_k(F_{\mathcal{A}^\prime}^{l-1}),\phi_v(F_{\mathcal{A}^\prime}^{l-1}),\mathbf{r},\mathbf{I}). \label{eq:supp_attn}
\end{align}
In Equation~\ref{eq:pos_enc}, $\text{{\fontfamily{pcr}\selectfont PE}}$ is the Fourier embedding layers. In Equation~\ref{eq:supp_attn}, $(i, j) \in \mathbf{I}$ and $\mathbf{I} \in \mathbb{N}^{K \times 2}$ indicates the directed or symmetric index set of total involved $K$ context-query token pairs which are determined through distance thresholds $r^{\text{q}\leftarrow\text{c}}$ (visible range centered on query token specified in Table~\ref{tab:supp_model}) with the upper limitation of number of total context tokens $N_c$.
While $\mathbf{r}\in\mathbb{R}^{K\times D}$ denotes the stacked positional encodings of total $K$ pairs. And $\phi$ represents the projection function to query, key and value, $l$ reflects the index of attention layer.


\vspace{2mm}
\mypar{Decoding Pose Token.} As described in Sec.~\ref{sec:interleave} and Fig.~\ref{fig:model}, we sample new pose token from the categorical distributions from the updated query feature $f_{a_0}$ at each autoregressive step, simultaneously with the control token. We detail the decoding process of the pose tokens:

we formulate the pose head as the sequentially connected position head and heading head, in this case we decode pose token through two separate steps.
Given an agent query $a_0$, we attach it to the ego agent to attend to environments with position-awareness, and produce the scene generation feature $q^{\prime}_{a_0}$.
We then first input it to position head to get the distribution over the position tokens $V_\text{pos}$, from which we get the position token of the new agent candidate through top-K sampling.
Secondly, we get the updated agent query $a^{\prime}_0$ by locating it at the accurate position which is translated from the position token, with its heading same as the one of the ego agent.
And again let $a^{\prime}_0$ to attend to its environment and output refined feature $q^{\prime}_{a^{\prime}_0}$, then send it to the heading head to get another probability distribution over the heading tokens $V_\text{head}$. The headings of the new agents are determined by sample the token with the highest probability.
In step of decoding headings of new agents, only Agent-Agent Attention and Map-Agent Attention will be employed, since we consider the headings are highly related to the surrounding agents and maps, but not the occupancy grids.

Note that, in Map-Agent Attention Layers, we use different visible range when decoding position tokens and heading tokens. For position tokens, we use an $r=75m$ while for heading tokens, $r=10m$ (also specified in Table~\ref{tab:supp_model}). We autoregressively repeat such position-heading decoding step which is controlled by control tokens.

\vspace{2mm}
\mypar{Decoding Attributes.} As mentioned in Eq.~\ref{eq:loss}, for those newly-inserted agents, we also calculate the losses of their shapes and types, since these attributes also have inherent relationship with their initial pose. We directly predict the continuous shapes value ($l,h,w$) through 3-layer MLP head. To get the types, we predict the categorical distributions over 3 defined types in WOMD (vehicle, cyclist and pedestrian) and take top-1 sampling.

\section{Token Details}
\label{sec:supp_token}

In this section, we have more details regarding the design of our tokenization mechanisms, especially the control tokens, and the formulations of GT tokens used for training. 

\vspace{2mm}
\mypar{Temporal Tokenization.} Regarding the tokens associated with temporal transitions, \textit{e.g.}, motion tokens, control tokens, we first tokenize the time axis at time span of $\delta=5$ step ($0.5\,\text{s}$ at $10\,\text{FPS}$), as Eq.~\ref{eq:tokenize_time}.
Accordingly, a real log from WOMD~\cite{montali2023waymo} ($n=91$ steps for $9.1\,\text{s}$) results in $N=\lfloor n/\delta\rfloor=18$ discrete tokens.
\begin{equation}
\label{eq:tokenize_time}
\text{Tokenize}(\{\mathbf{x}_0,\mathbf{x}_1,\cdots,\mathbf{x}_{n-1}\})=\{\hat{\mathbf{x}}_0,\hat{\mathbf{x}}_1,\cdots,\hat{\mathbf{x}}_{N-1}\},
\end{equation}
 where $\mathbf{x}$ denotes features with temporal transitions, \textit{e.g.}, motions ${x}^\text{m}$, validities ${x}^\text{v}$.
 Importantly, the value of each $\hat{\mathbf{x}}$ is defined based on different rules.
 For motions, each $\hat{{x}}_k^\text{m}$ is aggregated motions vector ${x}_{\delta k\,:\,\delta(k+1)}^\text{m}$ over the $k$-th time segment, as mentioned in Sec.~\ref{sec:tokens}.
 Furthermore, regarding the validity of agents, we consider both the starting and ending steps within its time segment: $\hat{{x}}_k^\text{v}$ is considered {\fontfamily{pcr}\selectfont True} if and only if both ${x}_{\delta k}^\text{v}$ and ${x}_{\delta(k+1)}^\text{v}$ are {\fontfamily{pcr}\selectfont True}, \textit{i.e.}, $\hat{{x}}_k^\text{v} = {x}_{\delta k}^\text{v} \land {x}_{\delta(k+1)}^\text{v}$, as a motion token is meaningful only when $\hat{\mathbf{x}}$ is valid.

\vspace{2mm}
\mypar{Control Tokens.} We aim to explain \textit{how to derive the control tokens from real logs.} Starting from the token-wise validity sequences (Eq.~\ref{eq:tokenize_time}), we define the \texttt{<ADD AGENT>} token as: 
\begin{equation}
\label{eq:add_agent}
\hat{{x}}_{k_1}^\text{c}
\;\; \text{when} \;\;
\hat{{x}}_{k_1}^\text{v} = \text{{\fontfamily{pcr}\selectfont True}} 
\;\text{ and }\; 
\forall\, 0 \leq i < {k_1}, \; \hat{{x}}_i^\text{v} = \text{{\fontfamily{pcr}\selectfont False}},
\end{equation}
and the \texttt{<REMOVE AGENT>} token $\hat{{x}}_{k_2}^{\text{c}}$ is symmetrically defined as the last valid token such that all subsequent ones are invalid, \textit{i.e.}, $\forall\, k_2 < i \leq N-1,\; \hat{{x}}_i^{\text{v}} = \text{{\fontfamily{pcr}\selectfont False}}$.
Thus, we set all tokens between \texttt{<ADD AGENT>} and \texttt{<REMOVE AGENT>}, $\hat{{x}}^\text{c}_k\,(k_1<k<k_2)$, as the \texttt{<KEEP AGENT>} token.

Notably, for two special cases: 1) we force the $\hat{{x}}_{k_1}^\text{c}$ with $k_1=0$ of Eq.~\ref{eq:add_agent} to be \texttt{<ADD AGENT>}; 2) for any \texttt{<REMOVE AGENT>} token $\hat{{x}}_{k_2}^\text{c}$ with $k_2=N-1$, we force it to be instead \texttt{<KEEP AGENT>}.

As we explained in Sec.~\ref{sec:interleave}, the tokens \texttt{<ADD AGENT>} and \texttt{<BEGIN MOTION>} only present when we examine the dynamic agent matrix column-wise.
Therefore, when we organize the tokens sequence according to the spatial layout (as opposed to Eq.~\ref{eq:tokenize_time}), \texttt{<BEGIN MOTION>} is defined as the next token after all the \texttt{<ADD AGENT>} tokens:
\begin{equation}
\hat{{x}}_{l}^\text{c}
\;\; \text{when} \;\;
\forall\, i < {l},\;
\hat{{x}}_{i}^\text{c} \;\text{ is }\; \texttt{<ADD AGENT>}.
\end{equation}
Moreover, for these \texttt{<ADD AGENT>} tokens of the GT spatial sequence, they are ordered according to the distances from the ego agent, as explained in Sec.~\ref{sec:train}.

\vspace{2mm}
\mypar{Empty Token.} Since we incorporate the invalid steps/agents, \textit{e.g.}, those with $\hat{x}^\text{v}=\text{{\fontfamily{pcr}\selectfont False}}$, into \METHODNAME model, we also involve the empty tokens shown in Figure~\ref{fig:model}, as part of dynamic agent matrix tensor $F_{\mathcal{A}^\prime}\in \mathbb{R}^{A\times T\times D}$. To build these invalid agent embeddings, we follow the similar methods of the agent query, but set their positions and headings to zeros.

Introducing these invalid values can bring unexpected noise within the modeling layers (in Sec.~\ref{sec:interleave}), specifically, interactions between tokens from different timesteps or different agents when any side of them has $\hat{x}^\text{v}=\text{{\fontfamily{pcr}\selectfont False}}$, which arise from two sources: 1) the construction of dynamic agent matrix tensor $F_{\mathcal{A}^\prime}$ (Eq.~\ref{eq:agent_embed}); 2) position-aware attention layers (Eq.~\ref{eq:rel_pos},~\ref{eq:pos_enc}). We address them through applying the rules:
\begin{subequations}
\label{eq:noise}
\begin{align}
    \hat{\mathbf{x}}_{\text{invalid}} - \hat{\mathbf{x}}_{\text{invalid}} &\leftarrow -\mathbf{z}_{\text{invalid}}, \\
    \hat{\mathbf{x}}_{\text{valid}} - \hat{\mathbf{x}}_{\text{invalid}} &\leftarrow \mathbf{z}_{\text{trans}}, \\
    \hat{\mathbf{x}}_{\text{invalid}} - \hat{\mathbf{x}}_{\text{valid}} &\leftarrow -\mathbf{z}_{\text{trans}},
\end{align}
\end{subequations}
where $\hat{\mathbf{x}}$ denotes tokens of various features, \textit{e.g.}, motion tokens $\hat{{x}}^\text{m}$, heading tokens $\hat{{x}}^\text{h}$, and $\hat{\mathbf{x}}_\text{invalid}$, $\hat{\mathbf{x}}_\text{valid}$ reflect their corresponding $\hat{{x}}^\text{v}$ are $\text{{\fontfamily{pcr}\selectfont False}}, \text{{\fontfamily{pcr}\selectfont True}}$, respectively.
We force these values to be our predefined constant values $\mathbf{z}$ to eliminate such noises due to the non-constant $\hat{\mathbf{x}}^\text{valid}$.
Since the model actually only needs the qualitative characteristic of the transition between invalid and valid states, rather than the specific quantitative values.
In our experiments, we set $\mathbf{z}_\text{trans}=1$ and $\mathbf{z}_\text{invalid}=-2$.
Without Equation~\ref{eq:noise}, \METHODNAME will suffer from the disruptive noises, preventing effective modeling of the control sequence.

\section{Training Details}
\label{sec:supp_train}
As we summarized in Sec.~\ref{sec:train}, we efficiently end-to-end train \METHODNAME model on multimodal token sequences.
Basically, we parallelly train temporal motion simulation and spatial scene generation as standard NTP task of each individual token modality, and perform interleaved autoregression in inference stage.
We break down the details of the each aspect for training process in this section.

\subsection{Temporal Simulation}

\vspace{2mm}
\mypar{Temporal Motions Training.}
We train on temporal motion tokens similar to prior works~\cite{nips24smart}, and additionally deal with the transition of before and after an agent is inserted or removed.
Given the temporal discrete tokens of one agent in Eq.~\ref{eq:tokenize_time}, we have their GT motion tokens $\{\hat{{x}}_k^\text{m}\}_{k=0}^{N-1}\subseteq\mathcal{V}_\text{motion}$, validities $\{\hat{{x}}_k^\text{v}\}_{k=0}^{N-1}\subseteq\{0,1\}$ ($0=\text{{\fontfamily{pcr}\selectfont False}}$, $1=\text{{\fontfamily{pcr}\selectfont True}}$), and, furthermore, the control tokens.
From the view of the temporal axis, \texttt{<ADD AGENT>} denotes the start of the sequence (BOS) while \texttt{<REMOVE AGENT>} denotes the end of the sequence (EOS).
Therefore, we refer to the states of the agent as its validities combined with BOS and EOS.

To supervise the motion tokens tensor $Y^\text{m}\in\mathbb{R}^{N}$ predicted by motion head, we derive the motion mask $M^\text{m}\in\mathbb{B}^N$ from the states\footnote{We omit the superscript of $M$ in this section when possible for simplicity.} (Note that masks $M$ are temporally aligned with prediction sequences, not groud-truth sequences).
Assuming the step of the BOS and EOS are $s_\text{BOS}$, $s_\text{EOS}$, then we have:
\begin{enumerate}[label=\arabic*) , leftmargin=*, itemsep=0pt]
  \item \(M_{s_{\text{BOS}}} = 1\): the step of BOS;
  \item \(M_{s_{\text{BOS}}+1} = x^{\text{v}}_{s_{\text{BOS}}+2}\) (with \(x^{\text{v}}_{s_{\text{BOS}}}=x^{\text{v}}_{s_{\text{BOS}}+1} = 1\)): the next step after BOS;
  \item $M_s=x^\text{v}_{s-1}\cdot x^\text{v}_s\cdot x^\text{v}_{s+1},\; \forall s,\;s_\text{BOS}+1<s<s_\text{EOS}$: the steps between the step after BOS and EOS (not included) only when the corresponding GT motions are valid.
\end{enumerate}
Otherwise, the steps not satisfy the conditions above have $M_s=0$, including the EOS.
Let $X^\text{m}\coloneq\{\hat{{x}}_k^\text{m}\}_{k=0}^{N-1}$, then the total loss for $N$ motion tokens is:
\begin{equation}
\label{eq:loss_motion_sequence}
    \mathcal{L}^\text{m}_{1:N}=\frac{1}{\left|\mathcal{I}\right|} \sum_{k \in \mathcal{I}} \text{{\fontfamily{pcr}\selectfont CE}}\big(Y^\text{m}_k,\;X^\text{m}_k\big),\;\mathcal{I} = \{ k \mid M_k^\text{m} = 1 \},
\end{equation}
where $\text{{\fontfamily{pcr}\selectfont CE}}$ is the CrossEntropy loss, as also described in Eq.~\ref{eq:loss_motion}.

\vspace{2mm}
\mypar{Temporal Controls Training.}
In the part of temporal simulation, we train on temporal control tokens $\{\hat{x}_k^\text{c}\}_{k=0}^{N-1}\subseteq\{\texttt{<NULL>}, \texttt{<KEEP AGENT>}, \texttt{<REMOVE AGENT>}\}$ similar to motion ones. We have described how to derive the control tokens \texttt{<KEEP AGENT>} and \texttt{<REMOVE AGENT>} from the validities in Sec.~\ref{sec:supp_token}.
And we have \texttt{<NULL>} as the placeholder token to indicate those steps without any control operations, which allows $X^\text{ct}$ to fully represent the entire GT temporal token sequence.

To supervise the control tokens tensor $Y^\text{ct}\in\mathbb{R}^N$ predicted by control head, we derive the temporal control mask $M^\text{ct}\in\mathbb{B}^N$ from the states. Here we have:
\begin{enumerate}[label=\arabic*) , leftmargin=*, itemsep=0pt]
  \item \(M_{s<s_{\text{BOS}}} = 0\): the steps before BOS (not included);
  \item \(M_{s_\text{BOS}} = 1\): the step of BOS;
  \item \(M_{s_{\text{BOS}}+1} = x^{\text{v}}_{s_{\text{BOS}}+2}\) (with \(x^{\text{v}}_{s_{\text{BOS}}}=x^{\text{v}}_{s_{\text{BOS}}+1} = 1\)): the next step after BOS;
  \item \(M_{s\geq s_{\text{EOS}}} = 0\): the steps after EOS (included);
  \item \(M_s=x^\text{v}_{s-1}\cdot x^\text{v}_s\cdot x^\text{v}_{s+1},\; \forall s,\;s_\text{BOS}+1<s<s_\text{EOS}\): the steps between the step after BOS and EOS (not included) only when the corresponding GT motions are valid.
\end{enumerate}
Then the total loss $\mathcal{L}^\text{ct}_{1:N}$ for the entire temporal control token sequence is calculated similar to Equation~\ref{eq:loss_motion_sequence} which takes $Y^\text{ct}$,$X^\text{ct}$,$M^\text{ct}$ as inputs.
Note that the steps with $M_s=1 (s_\text{BOS}<s<s_\text{EOS}-1)$ correspond to the \texttt{<KEEP AGENT>} tokens, while those with $M_s=1 (s=s_\text{EOS}-1)$ correspond to the \texttt{<REMOVE AGENT>} tokens. 
To alleviate the imbalance of these two control tokens, we set the label weights: $w(\texttt{<KEEP AGENT>})=0.1$ and $w(\texttt{<REMOVE AGENT>})=0.9$ when calculating the CrossEntropy Loss.

\subsection{Spatial Generation}

\vspace{2mm}
\mypar{Spatial Controls Training.}
For the spatial scene generation, we train on the control sequence $\{\hat{x}^\text{cs}_k\}_{k=0}^L\subseteq\{\texttt{<NULL>}, \texttt{<ADD AGENT>}, \texttt{<BEGIN MOTION>}\}$.
And $L$ denotes the total number of agents (including those existing and to be added) in a real log and we force $L=32$ in training process for saving memory.
We also use \texttt{<NULL>} as the placeholder for those agents without controls (\textit{e.g.}, existing and not to be added ones), allowing $X^\text{cs}$ to fully represent the entire GT spatial token sequence.

To formulate the spatial token sequence $X^\text{cs}$, we reorganize the all $L$ agents along the spatial axis, as explained in Sec.~\ref{sec:supp_token}, the first sequence (BOS) when scene generation begins, while \texttt{<BEGIN MOTION>} denotes sequence (EOS).
Hence, the tokens between BOS and EOS (not included) are all \texttt{<ADD AGENT>} tokens, and \texttt{<NULL>} only present after EOS which corresponds to those agents currently in motion simulation.

As a considerable number of the trailing tokens in $X^\text{ct}$ are \texttt{<NULL>} that cannot be trained, we further truncate the trailing part of $X^\text{ct}$---only consider the first $L^\prime=10$ tokens in training for more efficiency.
And the size of $X^\text{cs}$ in inference is \textit{scalable} since we train in an autoregressive way.

To supervise the control tokens tensor $Y^\text{cs}\in\mathbb{R}^{L^\prime}$ predicted by control head, we also have the mask $M^\text{ct}\in\mathbb{B}^{L^\prime}$ following:
\begin{enumerate}[label=\arabic*) , leftmargin=*, itemsep=0pt]
  \item \(M_{s_{\text{BOS}}} = 1\): the step of BOS;
  \item \(M_{s\geq s_{\text{EOS}}} = 0\): the steps after EOS (included);
  \item \(M_s=1,\; \forall s,\;s_\text{BOS}<s<s_\text{EOS}\): the steps between BOS and EOS (not included).
\end{enumerate}
Then the total loss $\mathcal{L}_{1:M^\prime}^\text{cs}$ for the spatial control tokens is obtained similar to Equation~\ref{eq:loss_motion_sequence}.
We also have label weights: $w(\texttt{<ADD AGENT>})=0.1$ and $w(\texttt{<BEGIN MOTION>})=0.9$ to deal with the class imbalance.

\begin{table*}[ht]
\centering
\renewcommand{\arraystretch}{1.0}
\caption{Ablation study of \METHODNAME on long-term traffic simulation ($\uparrow$). \checkmark\xspace indicates remaining unchanged as in Table.~\ref{table:newmetric}.}
\begin{threeparttable}
\begin{tabular}{c c c|c c c c c}
    \toprule[1pt]
     Cont. token & Pos. token & Head. token & Composite & Kinematic & Interactive & Map-based & Placement-based \\
    \midrule[0.5pt]
      & \checkmark & \checkmark & 0.6328 & 0.5493 & 0.7043 & 0.7961 & 0.4513\tnote{*} \\
     \checkmark &  & \checkmark & 0.6564 & 0.5580 & 0.7768 & 0.8077 & 0.4378 \\
     \checkmark & \checkmark & & 0.6509 & 0.5866 & 0.7276 & 0.8107 & 0.4445 \\
     & & \checkmark & 0.6297 & 0.5422 & 0.6962 & 0.7939 & 0.4499 \\
     \checkmark & \checkmark & \checkmark & 0.6674 & 0.5921 & 0.7688 & 0.8003 & 0.4503 \\
    \bottomrule[1pt]
\end{tabular}
\begin{tablenotes}
\footnotesize
\item[*] We take the heuristic approach to remove the agents.
\end{tablenotes}
\end{threeparttable}
\label{table:ablation}
\vspace{-7pt}
\end{table*}

\vspace{2mm}
\mypar{Spatial Hybrid Attention.}
To model such spatial sequence $X^\text{ct}$ in an autoregressive manner, we adapt the causal attention mechanism (in Agent-Agent Attention layers) similar to the Temporal Attention layers.

Specifically, when predicting token $\hat{\mathbf{x}}_t$ (\textit{e.g.}, motion token $\hat{x}_t^\text{m}$, control token $\hat{x}_t^\text{c}$) in temporal simulation, it will only involve the history tokens $\{\hat{\mathbf{x}}_{t-\tau}\}_{\tau=1}^{t_\text{w}}$ within the time window as the context (as Eq.~\ref{eq:temp_attn}).
Therefore, given the spatial token sequence $X^\text{cs}\in\mathbb{R}^{L^\prime}$ and there exist $A^\prime$ agents already in motion simulations, we construct a hybrid mask $M_\text{hybrid}^\text{ct}\in\mathbb{B}^{L^\prime \times (A^\prime+L^\prime)}$ which consists of two parts:
\begin{enumerate}[label=\arabic*) , leftmargin=*, itemsep=0pt]
  \item For those $A^\prime$ agents that already exist, they will not be masked out: $M_\text{hybrid}^\text{ct}[1\!:\!L^\prime,\; 1\!:\!A] = \mathbf{0}^{L^\prime \times A}$.
  \item For those $L^\prime$ agents to be predicted (may not all correspond to \texttt{<ADD AGENT>}), we have $M_\text{hybrid}^\text{ct}[1\!:\!L^\prime,\; A\!+\!1\!:\!A\!+\!L^\prime]$ to be a standard causal mask to exclude the futures in attention layers.
\end{enumerate}
We can wite it as:
\begin{equation}
    M_\text{hybrid}^\text{cs}[i,j]=
    \begin{cases}
    0, & \text{if } j \le i \text{ or } j < A^\prime \\
    -\infty, & \text{otherwise}
    \end{cases},
\end{equation}
where $i\in [1,L^\prime],\;j\in [1, A^\prime+L^\prime]$. Note that the ultimate context which query features attend to is determined by jointly applying $M^\text{cs}_\text{hybrid}$ and other possible masks (\textit{e.g.}, from the visible range).

In this way, we train on spatial token sequence with smaller context length $L^\prime=10$ while extend it to a scalable number ($>10$) with an upper limit of 128.


\section{Additional Results}
\label{sec:supp_more_results}

\subsection{Long-term Traffic Simulation}
\label{sec:supp_long_term}

\begin{figure*}
    \centering
    \includegraphics[width=\textwidth]{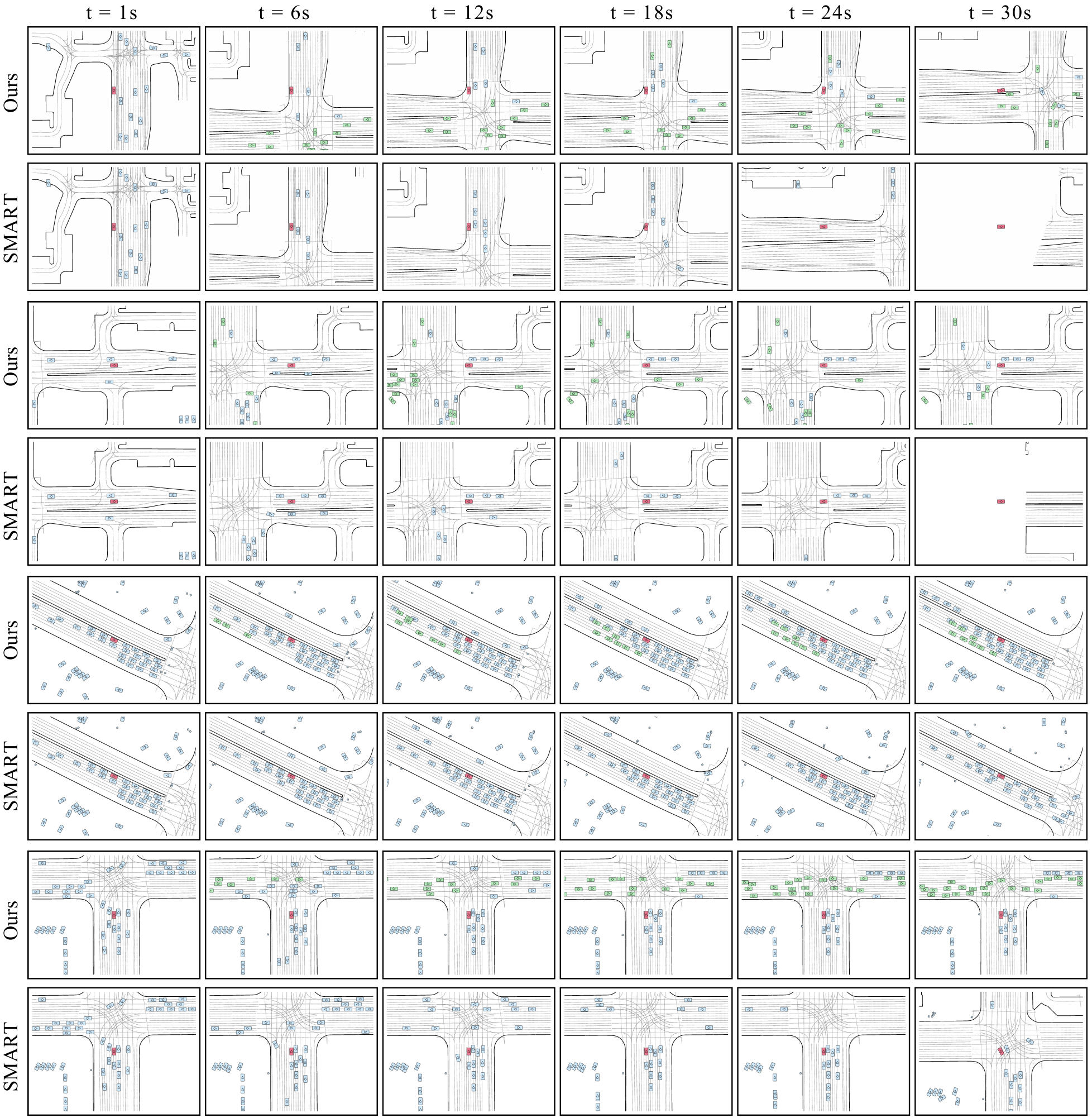}
    \caption{More qualitative comparison results \#1.}
    \label{fig:more_rollouts}
\end{figure*}

\begin{figure*}
    \centering
    \includegraphics[width=\textwidth]{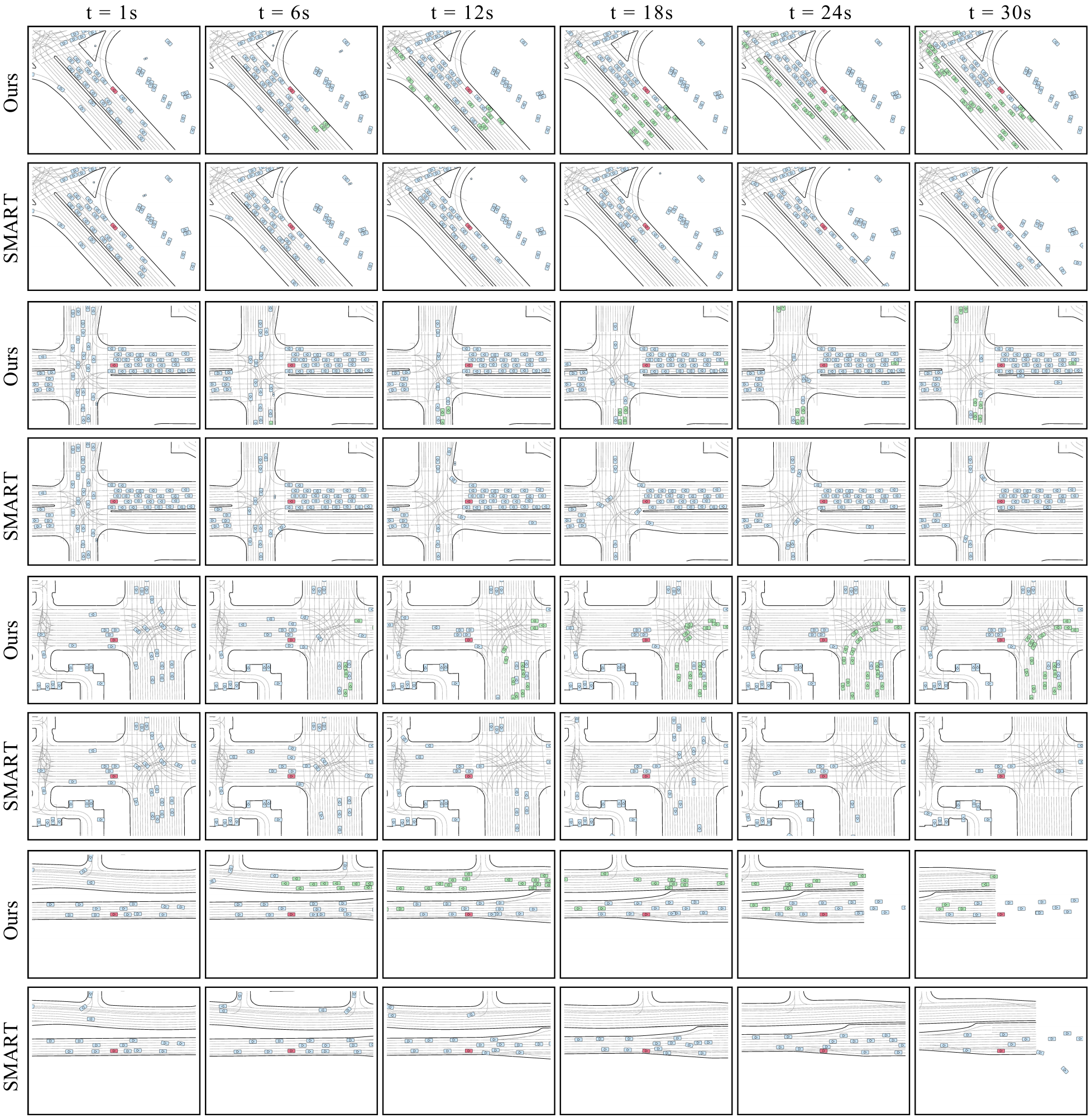}
    \caption{More qualitative comparison results \#2.}
    \label{fig:more_rollouts_2}
\end{figure*}

\begin{figure*}
    \centering
    \includegraphics[width=\textwidth]{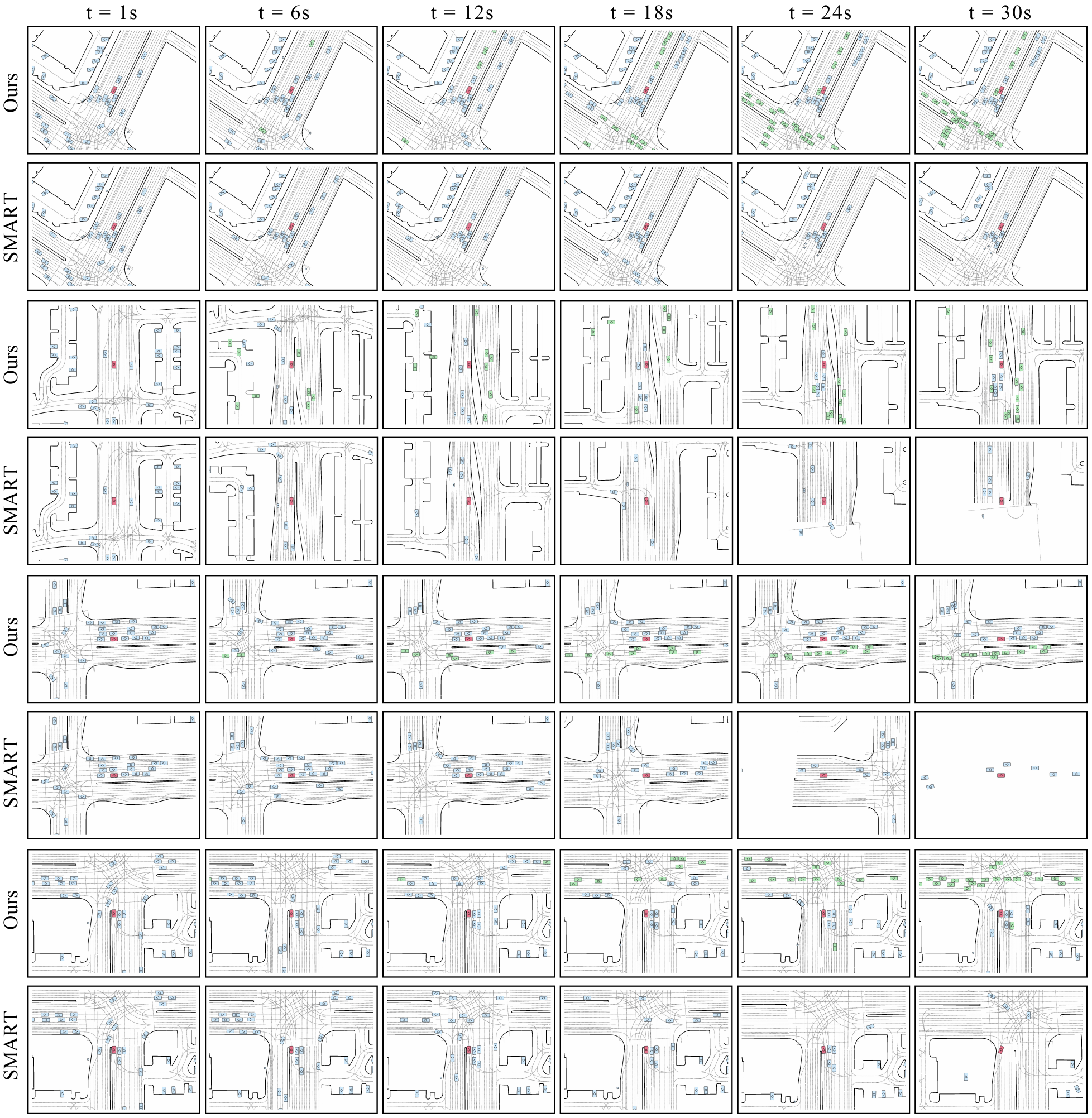}
    \caption{More qualitative comparison results \#3.}
    \label{fig:more_rollouts_3}
\end{figure*}

\vspace{2mm}
\mypar{More Qualitative Results.} We show more qualitative comparison results of \METHODNAME and the baselines [24,32] in Fig.~\ref{fig:more_rollouts},~\ref{fig:more_rollouts_2},and ~\ref{fig:more_rollouts_3}. In these scenario, we again demonstrate the strengths of our approach. As the ego agent travels for away from the initial locations, new agents appear in our examples while maintaining a great realism, seamlessly continuing the interaction process. This indicates that \METHODNAME can effectively one of the major challenges of \TASKNAME task.

\begin{figure}[t]
    \centering
    \includegraphics[width=0.8\columnwidth]{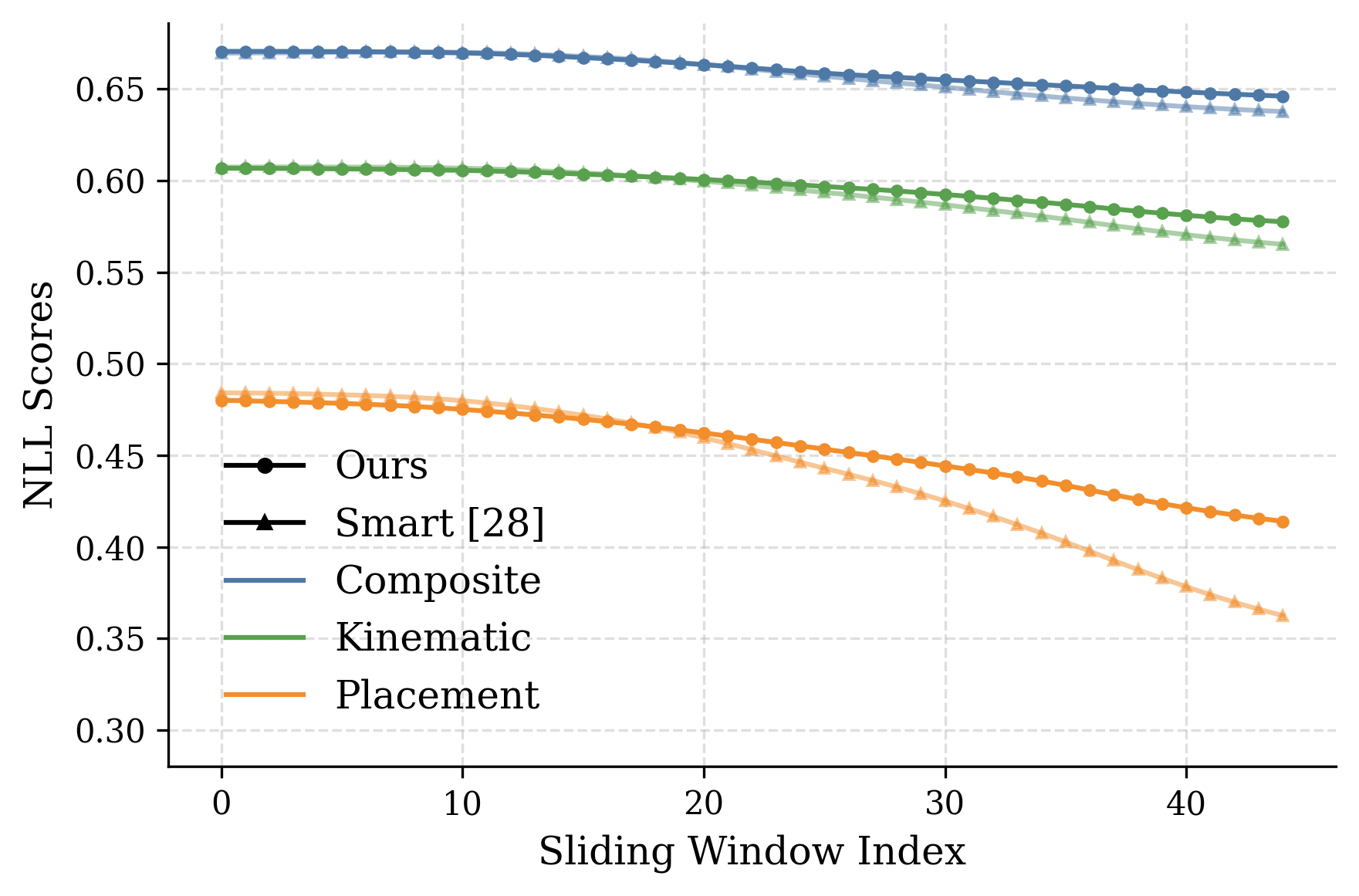}
    \caption{Metrics (adapted WOSAC) curve of \METHODNAME against SMART~\cite{nips24smart} over the 30s long-term simulation rollouts.}
    \label{fig:supp_curve}
    \vspace{-15pt}
\end{figure}

\vspace{2mm}
\mypar{Metrics Curves.}
We further investigate how simulation realism evolves over the duration of long-term rollouts by plotting metric scores for each sliding window index in Figure~\ref{fig:supp_curve}. 
Specifically, we show the evolution of three WOSAC metric components (Composite, Kinematic, and Placement-based) for both our method and SMART [24].

As expected, we observe a gradual decrease in realism scores across all methods, reflecting the increasing difficulty of maintaining realism over extended simulation periods. 
However, our method consistently outperforms SMART by exhibiting a notably slower decline in all metrics, particularly in the \textit{placement-based} component.
This significant improvement highlights the effectiveness of our proposed interleaved scene generation and motion simulation approach, enabling sustained realism by dynamically handling agent insertions and removals. 
These quantitative results strongly align with our qualitative observations, further emphasizing the importance of explicitly modeling agent placement and removal to achieve realistic long-term traffic simulations.

\subsection{Ablation Study}
\label{sec:supp_ablation}
We conduct various ablation studies to validate our methods. We ablate the impact of designed control tokens, position tokens and heading tokens on our task. Due to the high cost of local evaluation, following~\cite{zhang2024catk}, we use 5\% (2204 out of $\sim$44K scenarios) of the validation split in this part.

\vspace{2mm}
\mypar{Effect of Control Token.} As discussed in Sec.\ref{sec:tokens}, we introduce the control tokens to determine the spatial scene generation sequence.
The baselines, to some extent, can be regarded as versions without the \texttt{<ADD AGENT>} token, and naturally, the \texttt{<REMOVE AGENT>} token is also absent. To fully validate the control tokens, we additionally conduct long-term rollout tests while retaining the \texttt{<ADD AGENT>} token but removing the \texttt{<REMOVE AGENT>} token.

Note that we use the heuristic approach to remove agents with distances exceed the $R$, for two reasons: 1) Adding agents without removing any results in an unrealistic scenario that would not naturally exist; and 2) to ensure a fairer comparison. As shown in Table~\ref{table:ablation}, removing \texttt{<REMOVE AGENT>} token in long-term rollout severely degrades the kinematic and interactive metrics. Unsurprisingly, as the continuously increasing number of agents over time significantly impacts bother their motion states and internal interactions.

\vspace{2mm}
\mypar{Effect of Position Token.} We take position tokens to efficiently capture the environment information of local regions, which are ultimately aggregated into the agent query in spatial scene generation. We have an ablation experiment by completely removing the position tokens along with its token embedding, and we instead directly predict the $(x,y)$ locations of agents. The results reveal that the position tokens help the model better address the placement-related issues, as grids simplify the search space. Additionally, through position tokens embedding, the agent query can more efficiently perceive the spatial distribution of the environment.

\vspace{2mm}
\mypar{Effect of Heading Token.} The initialization of newly-entered agents' poses, is essential to the their subsequent motions and further the interactions with others. Similarly, we validate replacing the head token prediction with the direct prediction of continuous angle values. The results at table.~\ref{table:ablation} also reflects that heading tokens can slightly improve the interactive and placement-based performance.




\section{Limitations and Future Direction}
\label{sec:fail}

Although \METHODNAME has achieved promising results on \TASKNAME, our method is limited in some aspects, as briefly described in Sec.~\ref{sec:conclusion}. In this section, we have detailed discussions on these terms.

\begin{figure}[t]
    \centering
    \includegraphics[width=\columnwidth]{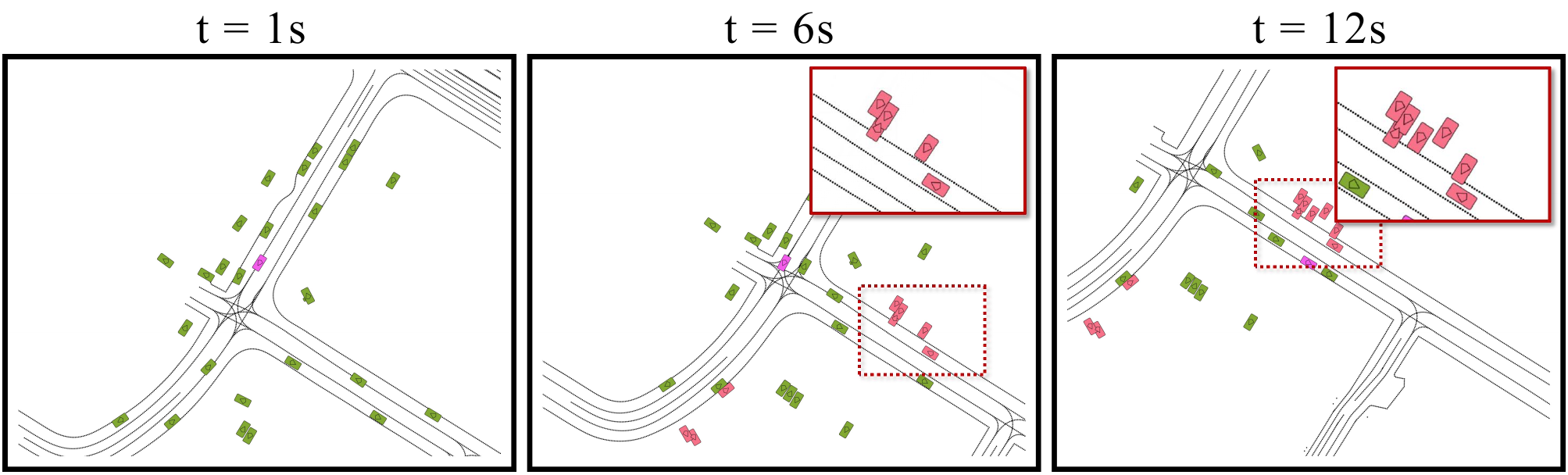}
    \caption{Failure case \#1: newly-entered agents appear unreasonably on the boundary of the road.}
    \label{fig:failure_1}
\end{figure}

\vspace{2mm}
\mypar{Failure Cases.} We have some failure cases existed:

\textit{(1) Unreasonable inserted agents.} In some examples, our method may have unreasonable newly entered agents in traffic scenario. As shown in Fig.~\ref{fig:failure_1}, at $t=6\,\text{s}$ and $t=12\,\text{s}$, the agents highlighted by red boxes occupy the road boundaries, which is unrealistic in real-world scenarios. It indicates that our method lacks sufficient control at such a fine-grained level. Notably, we did not impose any explicit constraints on this kind of cases, such as regularization losses.

\begin{figure}[t]
    \centering
    \includegraphics[width=\columnwidth]{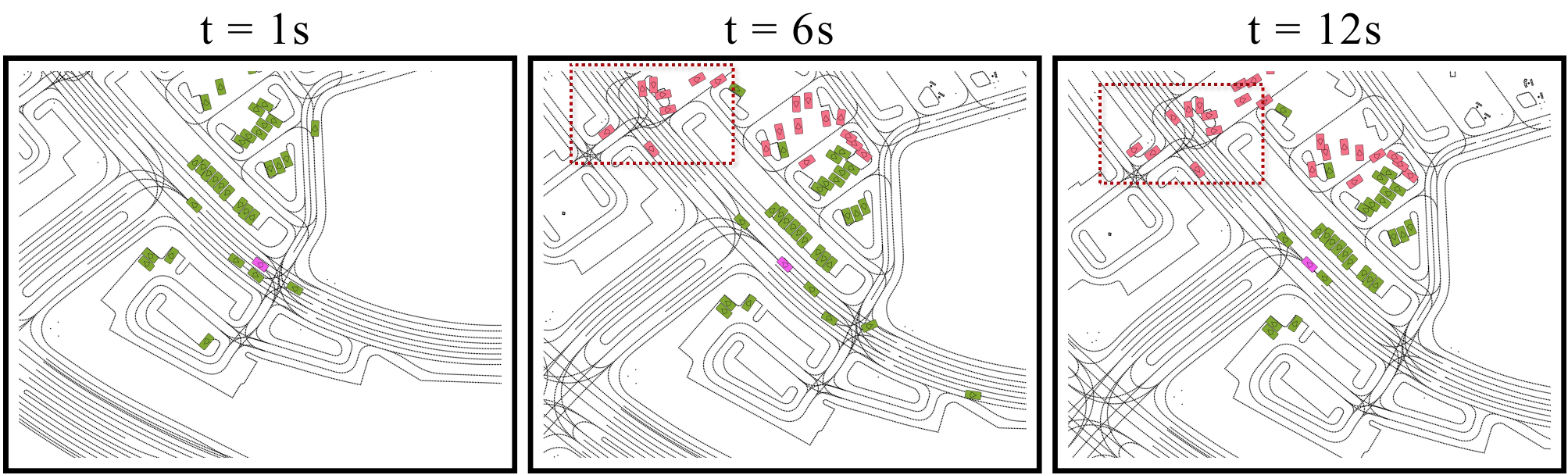}
    \caption{Failure case \#2: In the region with complex map structure (highlighted by red box), newly-entered agents fail to correctly infer their initial velocity (or motion) and remain stationary, which is unrealistic.}
    \label{fig:failure_2}
\end{figure}

\textit{(2) Incorrect initial motion inferring.} During the spatial sequence prediction, \METHODNAME first observes overall traffic scenario before placing new agents in potential locations. When these agents are located in a complex road situation, they may fail to accurately infer their initial velocity (or motion) for the subsequent rollout. For example, as shown in Fig.~\ref{fig:failure_2}, some new agents incorrectly remain stationary on the driving lanes, which also impacts the motions of other agents, ultimately reducing the realism.

\textit{(3) Agents flickering modeling.} According to our observations, the phenomenon of agents ``flickering'' is prevalent in real-world data, particularly in regions farther from the ego agent.
And it arises due to the instability of the ego agent's remote perception. In our work, we handle these flickering agents in two ways: first, we discard agents with a presence duration shorter than $0.5\,\text{s}$; second, we change the flickering frequency, which can be attributed to the resolution of discretization on time axis (as introduced in Sec.~\ref{sec:supp_token}).
Specifically, the minimum temporal granularity that each token can represent is $0.5s$.
As a result, \METHODNAME inherently struggles to generate highly realistic agents exhibiting flickering behavior.

A potential solution is to introduce another format of tokens $\mathcal{V}_\text{validity}$ which reflects the frame-wise validity within even one $0.5\,\text{s}$ token, and make \METHODNAME learn it in temporal simulation. Given that each $0.5\,\text{s}$ segment contains 5 valid timesteps, with each step has 2 possible states (invalid, and valid), we can derive the vocabulary size: $2^5=32$. But the task becomes more complex under such conditions. We leave this as a future direction.

\vspace{2mm}
\mypar{Future Directions.} In addition to addressing the limitations discussed, some other potential interesting improvements may be as below:


\textit{(1) Long context understanding and learning.} While \METHODNAME effectively addresses the challenge faced by Long-term Sim Agent---modeling the insertion and deletion of agents interleaved with their motions in continuously evolving traffic scenarios to maintain high realism over extended rollout durations---we believe that another critical bottleneck for even longer horizons is the accumulation of errors over the rollout process. In our work, we adhere to a unified next-token prediction paradigm for end-to-end training, with reference to a historical information constrained by the temporal length. Some hard cases in a long-long-term rollout may fail: in a busy intersection where the lateral traffic passes, followed by the longitudinal traffic while other different lanes remain open, the execution intervals of different actions can be significantly long, and the rules can be greatly complex. Some works~\cite{zhang2024catk,tan2024promptable} utilize closed-loop training or finetuning for improvements, which also cannot be a solution. Thus, it remains an other challenge.

\textit{(2) Driving Map Generation.} The duration of long-term rollout controlled by \METHODNAME is significantly constrained by the size of the map region----without this limitation, it would be possible to extend the rollout even further. Therefore, integrating the map generation would be a substantial improvement, enabling longer and more flexible simulations. Some concurrent works, such as GPD-1~\cite{xie2024gpd1}, have some progress in this point, making it a promising direction for near future.

\section{License}

The Waymo Open Motion Dataset (WOMD)~\cite{montali2023waymo} we used in our work is licensed under Waymo Dataset License Agreement for Non-Commercial Use\footnote{\href{https://waymo.com/open/terms/}{https://waymo.com/open/terms/}}.

The implementations of official WOSAC metrics\footnote{\href{https://github.com/waymo-research/waymo-open-dataset/}{https://github.com/waymo-research/waymo-open-dataset/}} based on which we develop our extended metrics are licensed under the Apache License, Version 2.0.






\end{document}